\documentclass[conference]{IEEEconf}

\usepackage{cite}      
\input epsf
\usepackage{graphicx}
\usepackage{booktabs} % To thicken table lines
\usepackage{caption}

\usepackage{subfig}

\usepackage{alphalph}
\usepackage{tikz}
\usepackage{multirow}
\usepackage{amsmath}   
\usepackage[T1]{fontenc}
\usepackage{amsfonts}

\begin{document}
	
	% paper title
	%\title{Submission Format for IPVC-CyberSec21 (Title in 24-point Times font)}
	% If the \LARGE is deleted, the title font defaults to  24-point.
	% Actually, 
	% the \LARGE sets the title at 17 pt, which is close enough to 18-point.
	%+++++++++++++++++++++++++++++++++++++++++++
	\title{\textbf{\Large Evaluating the Robustness of Deep Reinforcement Learning for Autonomous Policies in a Multi-agent Urban Driving Environment\\}}
%	\author{Blind Review}
	\author{\IEEEauthorblockN{Aizaz Sharif}
	\IEEEauthorblockA{\textit{Simula Research Laboratory} \\
		Oslo, Norway \\
		aizaz@simula.no}
	\and
	\IEEEauthorblockN{Dusica Marijan}
	\IEEEauthorblockA{\textit{Simula Research Laboratory} \\
		Oslo, Norway \\
		dusica@simula.no}
	
}
	%+++++++++++++++++++++++++++++++++++++++++++
	
	% use only for invited papers
	%\specialpapernotice{(Invited Paper)}
	
	% make the title area
	\maketitle
	\begin{abstract}
  \textit{Background}: Deep reinforcement learning is actively used for training autonomous car policies in a simulated driving environment. Due to the large availability of various reinforcement learning algorithms and the lack of their systematic comparison across different driving scenarios, we are unsure of which ones are more effective for training autonomous car software in single-agent as well as multi-agent driving environments. \textit{Aims}: A benchmarking framework for the comparison of deep reinforcement learning in a vision-based autonomous driving will open up the possibilities for training better autonomous car driving policies. \textit{Method}: To address these challenges, we provide an open and reusable benchmarking framework for systematic evaluation and comparative analysis of deep reinforcement learning algorithms for autonomous driving in a single- and multi-agent environment. Using the framework, we perform a comparative study of four discrete and two continuous action space deep reinforcement learning algorithms. We also propose a comprehensive multi-objective reward function designed for the evaluation of deep reinforcement learning-based autonomous driving agents. We run the experiments in a vision-only high-fidelity urban driving simulated environments. \textit{Results}: The results indicate that only some of the deep reinforcement learning algorithms perform consistently better across single and multi-agent scenarios when trained in various multi-agent-only environment settings. For example, A3C- and TD3-based autonomous cars perform comparatively better in terms of more robust actions and minimal driving errors in both single and multi-agent scenarios. \textit{Conclusions}: We conclude that different deep reinforcement learning algorithms exhibit different driving and testing performance in different scenarios, which underlines the need for their systematic comparative analysis. The benchmarking framework proposed in this paper facilitates such a comparison.
		
		%	For example, A3C- and TD3-based autonomous cars perform comparatively better in terms of more robust actions and minimal driving errors in both single and multi-agent scenarios.
		%only some of the DRL algorithms perform consistently better across single and multi-agent scenarios when trained in a multi-agent only setting. We also conclude that there can be more than one RL algorithm choice for training adversarial agents that can be used in AD driving research. We further conclude that 

	\end{abstract}
	\IEEEoverridecommandlockouts
	% no keywords
	
	% For peer review papers, you can put extra information on the cover
	% page as needed:
	% \begin{center} \bfseries EDICS Category: 3-BBND \end{center}
	%
	% for peerreview papers, inserts a page break and creates the second title.
	% Will be ignored for other modes.
	\IEEEpeerreviewmaketitle
	
 \section{Introduction}\label{sec:Introduction}
Autonomous cars (ACs) are complex decision-making systems that are unfortunately prone to errors \cite{Garcia}. They commonly use machine/deep learning algorithms as part of decision-making software, which is known to be difficult to validate \cite{Riccio, MLTest}. Therefore, ACs need comprehensive training and evaluation in order to minimize risk to the public. For the same reason, autonomous driving (AD) research is nowadays performed within simulated driving environments (also used by the state-of-the-art industrial solutions like Tesla~\cite{tesla} and Comma ai~\cite{comma}), as they provide flexibility for testing and validating AD without posing any danger to the real world. However, we observe three specific challenges in training and validating ACs in the existing simulation environments.

First, while the majority of AD research is focused on using deep reinforcement learning (DRL) for training ACs in a simulated urban driving scenario~\cite{PPO}\cite{A3C2}\cite{DQN2}\cite{DDPG}\cite{DDPG2}, there is a lack of comparison among DRL algorithms for vision-based urban driving scenarios. Having such a benchmark of commonly used DRL algorithms can be useful for understanding why some algorithms perform worse than others in specific driving scenarios, which can lead to the improvements of the state-of-art DRL algorithms for AD.

Second, the majority of existing research trains ACs as non-communicating and independent single intelligent agents~\cite{pmlr-v155-zhou21a}\cite{doi:10.1073/pnas.1820676116}, therefore treating the ACs as a single-agent driving problem. However, in the near future AD will be a complex multi-agent problem~\cite{R13}. By bringing more than one AC into a multi-agent environment we can evaluate how a non-stationary driving scenario affects AC’s control decisions when interacting with other ACs. Existing comparative analyses of RL algorithms for AD are still limited to single-agent driving environments~\cite{benchmark}\cite{Vinitsky2018}, and there is no systematic study performed yet on which DRL models work best for AD in a multi-agent environment.

Third, a vast portion of existing research is actively focused on testing ACs trained on vision-based end-to-end systems. While designing a reward function is one of the key requirements in learning an efficient and robust DRL driving agent, we lack evidence of a detailed multi-objective-based reward function that is utilized within independent multi-agent DRL algorithms. There is ongoing research on designing a better reward function~\cite{huang2022efficient}\cite{chen2019model} which is highly required for better autonomous driving, but so far they are still focused on single-agent driving scenarios. Furthermore, in order to compare which DRL works the best in single and multi-agent-focused scenarios, their reward functions must also be defined in a comprehensive way in order to perform a fair comparison in terms of robustness. As DRL heavily relies on the design of the reward function for active exploration and learning, it is necessary to consider various objectives within a reward function in order to observe and evaluate which objectives are successfully achieved by each DRL-based AC agent within different driving settings. We, therefore, also lack studies on the performance of DRL algorithms that are trying to achieve multiple goals using the reward function.

To address these three challenges, in this paper we provide an open and reusable end-to-end benchmarking framework for DRL algorithms in a complex urban multi-agent AD environment. The framework enables us to bring competitive and independent driving agents in both discrete and continuous action space environments. We also propose a comprehensive multi-objective reward function designed for the evaluation of simulation-based learning of DRL AC agents. We validate the framework by performing a comprehensive comparative study of the robustness of DRL algorithms for urban driving policies in both single-agent and multi-agent AD scenarios.

	%One of the ways to test their control behavior is using adversarial RL (ARL). In fact, DRL is proved to be vulnerable multiple times against adversarial attacks~\cite{R7}, and one of the solutions to this problem is to train the adversary as a separate physical driving model using DRL~\cite{R4}. ARL can be used to not only find failure scenarios in ACs but also to improve their driving policies through retraining~\cite{R8}. However, at the moment, we lack the support for training and systematically evaluating ARL algorithms to decide which ones are more effective in creating better adversarial AD agents able to find errors in ACs.

 The key contributions in this paper are:
\begin{enumerate}
	\item We provide an end-to-end benchmarking framework for the systematic evaluation of DRL-based AC driving policies in complex vision-based urban driving environments.
	\item We propose a multi-objective reward function designed for the evaluation of simulation-based learning of DRL AC agents.
	\item The framework supports training and validating AC's driving policies in both single- and multi-agent driving scenarios.
	\item The framework enables evaluating the effectiveness of AC driving agents in different environment configurations.
	\item Using the framework, we perform a comprehensive comparative study of four discrete and two continuous action space DRL algorithms for AD in vision-only high fidelity urban driving simulated environments.
	\item Drawing from the study results, we suggest some research directions on improving the robustness of DRL algorithms for AD.
	\item The implementation of our benchmarking framework, as well as all experimental results, are open and reusable, which supports the reproducibility of research in the AD domain.
\end{enumerate}

\section{Related Work}
In AD research, there are only a few benchmarks for evaluating the performance of RL-based AD models. Vinitsky \cite{Vinitsky2018} proposes a benchmark for DRL in mixed-autonomy traffic. While the benchmark involves four scenarios: the figure eight network, the merge network, the grid, and the bottleneck, it evaluates a limited number of reinforcement learning (RL) algorithms (two gradient-based and two gradient-free).
%In addition, the authors do not consider ARL in their work, although dealing with adversarial attacks is a great challenge in AD research.
In addition, the proposed benchmark is specific to connected AD research. Stang \cite{benchmark} proposes another benchmark for RL algorithms in a simulated AD environment. As a limitation, this benchmark focuses on a simple lane-tracking task, and furthermore, evaluates only off-policy RL algorithms. In contrast, our work evaluates both on-policy and off-policy algorithms and further allows for comparing the performance of DRL algorithms in a complex urban environment. As another limitation, both \cite{Vinitsky2018} and \cite{benchmark} only support DRL benchmarking for single-agent AD environments.

Li \cite{LiArxiv2021} introduces a driving simulation framework called MetaDrive and performs a benchmarking of RL algorithms for AD. While the authors use five different driving scenarios, they only evaluate two RL algorithms (PPO and SAC).
%The proposed work also lacks any research on dealing with adversarial attacks on driving policies.
The work could also benefit from using realistic visual rendering as provided by the CARLA framework in our work. Palanisamy~\cite{IMPALA} proposes a multi-agent urban driving framework in which one can train more than one AC. Using IMPALA, a connected AC policy is trained within the CARLA simulator. However, as a limitation, the work is restricted to connected AD problems only.

Furthermore, there are frameworks proposed for training and testing autonomous vehicles. For example, F1TENTH framework \cite{kelly20a} with three racing scenarios and baselines for testing and evaluating autonomous vehicles. However, the framework does not support dealing with ARL. Han \cite{9552860} proposes an off-road simulated environment for AD with realistic off-road scenes, such as mountains, deserts, snowy fields, and highlands. While realistic environments are useful for evaluating the generalization abilities of AD models, the work is limited to single-agent AD environments.

\section{Deep Reinforcement Learning for Autonomous Driving}\label{sec:Background}
Reinforcement learning is mainly modeled as a formulation of the Markov Decision Process (MDP), where the desired goal of the agents in a certain environment is to learn an optimal policy. This goal is achieved by maximizing cumulative reward after interacting with an environment. The MDP model consists of $M(S, A, P, R,\gamma)$, where $S$ is a set of agent's state and $A$ is a set of discrete or continuous actions. $R:S \times A \mapsto \mathbb{R}$ is a reward function value returned against every action $A$, whereas $\gamma$ is the discount rate applied to the future reward values. Lastly, the MDP model consists of a $P: S \times A \mapsto S$ as the transition probability which describes the stochastic probability distribution of the next state $s'$ $\epsilon$ $S$ given actions. Following the basics of the MDP model, the agent is dependent on the previous state only to make the next decision. Such a system obeys Markov property during RL control decisions.
%\subsection{Reinforcement Learning}\label{sec:RL}

Reinforcement learning has achieved great results over the past few years due to the advancements in DRL. In DRL, the MDP model is solved by using deep neural networks to learn weight parameters $\theta$ over the time span of environment exploration. DRL models consist of a policy $\pi$ which is responsible for taking an action given a state $\pi(a|s)$ and a value function $v_\pi s$ for estimating maximum reward given the current policy $s$ $\epsilon$ $S$. Traditional RL fails at solving high-dimensional state space problems and therefore deep neural networks enable the possibility of learning a function approximation over large input and action state spaces to solve complex problems. The policy and value functions in DRL are therefore learned using the defined deep net models to estimate future actions. In our work, the DRL algorithms we choose to benchmark are based on a model-free approach. In model-free RL, the policy $\pi$ and value function $v_\pi s$ are learned directly by interacting with the environment without taking model dynamics and the transition probability function of every state space into consideration. 

Next, we provide a brief description of the DRL models we use for the training and validation of AD in a simulated urban environment. These models are chosen on the basis of i) popularity in the DRL-based AD research community and ii) coverage of discrete as well as continuous action space (for multi-agent testing purposes).
\subsection{DRL Algorithms for Autonomous Driving}\label{sec:DeepRL}
\subsubsection{Discrete Action Space}
\hfill

\textbf{Proximal Policy Optimization (PPO)}:\label{sec:PPO}

PPO~\cite{R55} is a DRL algorithm, which also serves as one of the extensions to the policy gradient (PG) algorithms. Vanilla PG faces the problem of high gradient variance, and therefore PPO solves it by adding constraints like clipped surrogate objective and the KL penalty coefficient. Such improvements made PPO a very easy choice in the domain of DRL over the past few years.

In terms of vision-based AD, the authors in~\cite{PPO} use PPO as an RL algorithm to train a driving policy using synthetic simulated RGB images. The trained policy is transferred to real-world testing experiments for analyzing the perception and control perspective of the AC. Another authors in~\cite{PPO2} uses PPO for proposing a road detecting algorithm in an urban driving environment. They carry out experiments in a Udacity racing game simulator~\cite{udacity} as well as in a small OpenAI gym carracing-v0 environment~\cite{gym}.

	%\paragraph{\textbf{Advantage actor-critic (A2C)}}\label{sec:A2C}
	%A2C~\cite{conf_ndss17_a2c} is a synchronous version of the A3C RL algorithm. A2C is designed to solve problems associated with A3C regarding how individual actors can update the global parameters and thus it can lead to non-optimal solutions. In A2C, a coordinator is introduced in order to bring consistency to the actors so that all the actors can work with the same policy in the next iterations. 
	%
	%Jaafra proposed~\cite{A2C} to addresses an AD problem using CARLA autonomous simulator for training their DRL agent. Using an actor-critic architecture, the driving policy takes RGB images as input and learns to take action outputs within an urban driving environment.  
	
{\textbf{Asynchronous Advantage Actor-Critic (A3C)}:}\label{sec:A3C}

A3C~\cite{pmlr} is a well-known gradient descent-based RL algorithm. Following the on-policy technique, it focuses on using two neural networks - actor and critic. An actor is responsible for making actions while the critic aims for learning a value function. A3C takes this approach by keeping a global critic model while making multiple actor models for parallel training.  

A3C has been used in~\cite{A3C} for training an AC policy in a virtual environment that can work in a real-world situation as well. By using synthetic images as an input, the policy learns to drive in an urban scenario using the proposed DRL method. Another work by the authors in~\cite{A3C2} also uses A3C by combining RL with image semantic segmentation to train an AC driving policy. Using the TORCS~\cite{Torcs} racing simulator, their goal is to lower the gap between virtual and real models while training an RL agent. A3C is also used in~\cite{A3C3} where the authors created an end-to-end driving approach without detailed perception tasks on input images.  

\textbf{Importance Weighted Actor-Learner Architecture (IMPALA):}\label{sec:IMPALA}

IMPALA~\cite{impalaa} uses an actor-critic technique, but with the twist of decoupling actors. Following a V-trace off-policy approach, IMPALA's main objective is to scale up the DRL training capability by adding multiple independent actors. The actors, in this case, are meant to generate experience trajectories for the policy to learn, while the learner tries to optimize not only the policy but also the value function.

An author in~\cite{IMPALA} proposes a multi-agent urban driving framework in which one can train more than one AC. Using IMPALA, the author performs training on connected AC policy within the CARLA simulator.

{\textbf{Deep Q Networks (DQN):}}\label{sec:DQN}

DQN~\cite{Mnih} falls into the category of value-based methods, where the goal is to learn the policy function by going through the optimal action-value function. DQN uses off-policy and temporal differences (TD) to learn state-action pairs by collecting episodes as experience replay. Using the episodic data, samples from the replay memory are used in order to learn Q-values, as a neural network function approximation. DQN is one of the foundation models in the upbringing of DRL and there are many improved versions of DQN implemented by overcoming the existing flaws.  

DQN is used in~\cite{DQN} for training an AC in a Unity-based urban driving simulator. The authors use cameras and lasers as their input sensors for training the driving policies. Another work in~\cite{DQN2} uses DQN for first training a driving agent in simulation to test its navigation capabilities in both simulated and real-world driving situations. DQN is also utilized in~\cite{DQN3} for learning to steer a vehicle within a realistic physics simulator. Given the camera input feeds, their DQN-based agent is aiming for following lanes with minimal offroad steering errors. 

\subsubsection{Continuous Action Space}

\hfill

{\textbf{Deep Deterministic Policy Gradient (DDPG):}}\label{sec:DDPG}

When it comes to continuous action space, DDPG~\cite{ddpg_original} is the most widely used algorithm in DRL research. DDPG is a model-free and off-policy approach falling under the actor-critic algorithms. It extends DQN by learning a deterministic policy in a continuous action space using actor-critic framework.

DPG is extensively used in the field of training autonomous vehicles within simulated driving scenarions~\cite{DDPG}. Authors in~\cite{DDPG2} uses DDPG to construct a DRL model for learning to avoid collisions and steering. DDPG is also used in~\cite{DDPG3} for learning a lane following policy within driving simulation using continuous action space. 

{\textbf{Twin Delayed DDPG (TD3):}}\label{sec:TD3}

TD3~\cite{TD3_original} extends the idea of DDPG in continuous action space algorithms by tackling issues regarding overestimation of the Q-learning function found in the value-based and actor-critic methods. This results in both improving the learning speed and model performance of DDPG within a continuous action setting.

TD3 is extensively used in urban driving scenarios for training AC agents. As a model-free approach, TD3 is used in~\cite{TD31}\cite{TD32} for learning an AD policy within an urban simulated environment. The authors proposed a framework for learning complex driving scenarios using visual encoding to capture low-level latent variables. Another work in~\cite{TD33} also uses TD3 in to overcome the challenges of driving in an urban simulated environment.

\subsection{Multi-agent Autonomous Driving}\label{sec:Multiagent}
While introducing multi-agent AD agents, we need to consider an environment where agents do not have access to all the states at each time step. Such types of environments are found in the field of robotics and ACs where an agent is limited to the sensory information gathered by its hardware. Therefore, the existing MDP can be termed as a Partially Observable Markov decision process (POMDP)~\cite{R58}. Furthermore, the current formulation of POMDP can be reformulated as Partially Observable Stochastic Games (POSG)~\cite{R57} by defining a DRL control problem as a tuple $(I, S, A, O, P, R)$. In POSG, we can incorporate multi-agent scenarios using Markov Games~\cite{R56} where multiple agents are interacting with the environment. An actor $i$ $\epsilon$ $I$ receives its partial observations from a joint observation state $o_i$ $\epsilon$ $O_i$ at each time step $t$. Following the traditional MDP approach, each actor uses its learned policy function $\pi_i : O_i \mapsto A_i$ to perform actions $a_i$ $\epsilon$ $A_i$. As a return, each actor gets a desired reward value $r_i$ $\epsilon$ $R_i$.
	
	%\subsection{Adversarial Reinforcement Learning} \label{sec:ARL}
	%ARL is a new branch of RL where adversarial algorithms are trained such that they create perturbation attacks against a victim. The desired objective of ARL is to find failure outputs in the victim when the victim is exposed to such adversaries. The victim ACs are obtained from experimental evaluation as described in Section~\ref{sec:Step1} and~\ref{sec:Step2}. In our work, the adversarial agents are trained as physical driving agents, as opposed to victim ACs in a competitive urban driving environment. Such driving adversaries have no white box access to the input, output, or model weights of victim ACs. By following a zero-sum game strategy, they learn a policy that produces adversarial output actions that appear as natural observations for victim ACs, thus fooling them into errors.%, for gaining positive reward.

	\section{End-to-end benchmarking framework}\label{sec:Setup}
	In this section, we provide the details of the DRL benchmarking framework architecture as well as an explanation of the reward function, hyperparameters, and driving policies. The implementation of the framework is available at~\footnote{https://github.com/AizazSharif/Benchmarking-QRS-2022}.
	
	\subsection{Driving Policies}\label{sec:Policies}
	We divide our driving policies into $\pi_{AC}$, where $\pi_{AC}$ represents the policy of AC driving agents that are trained using one of the six selected DRL algorithms. Policies and their associated algorithm symbols are mentioned in Table~\ref{tab:drivingpolicies}. Their usage is thoroughly explained in Section~\ref{sec:Steps}.
	
	\begin{table}[htbp]
		\caption{Driving Policies and their associated symbols for ACs.} \label{tab:drivingpolicies} 
		
		\begin{center}
			\resizebox{!}{1.6cm}{
				\begin{tabular}{llll}
					\toprule
					\textbf{DRL Algorithm} & \textbf{Notation for AC Policy}  \\ \midrule
					
					PPO	& $\pi_{PPO}$     \\ 
					A3C	& $\pi_{A3C}$    \\ 
					IMPALA	& $\pi_{IMPALA}$   \\ 
					DQN	& $\pi_{DQN}$     \\ \midrule
					DDPG	& $\pi_{DDPG}$    \\ 
					TD3	& $\pi_{TD3}$    \\ \bottomrule
				\end{tabular}
			}
		\end{center}

	\end{table}

 \subsection{Deep Neural Network Model}\label{sec:NN}
\begin{figure*}[!htbp]
	\centering
	
	\includegraphics[width = 0.92\textwidth,height=0.25\textheight]{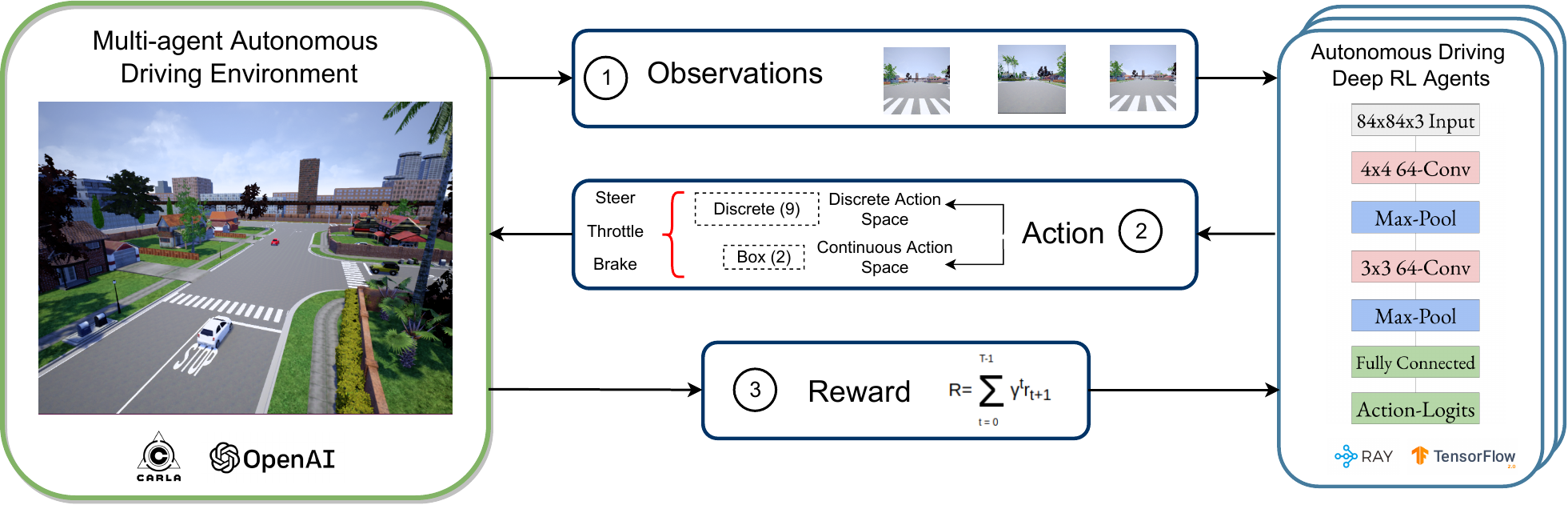} \caption{End-to-end DRL benchmarking framework for AC agents. An agent receives an input observation image of 84x84x3 which is passed to a DRL model. Actions are selected at the output layer of every agent and are performed in the next time step of the simulation in order to obtain a reward and a new observation state.
	}
	\label{fig:Architecture}
\end{figure*}

A pictorial description of the DRL benchmarking framework can be seen in Figure~\ref{fig:Architecture}. Each driving agent as AC receives partial input state observation of 84x84x3 dimension images through the front camera sensors. Cameras are mounted as part of the driving agents, and during each time step of the simulation environment, cameras capture the input state observations which serve as an input layer to the DRL model. The input layer is then passed to the convolutions and connected to hidden layers for extracting important features before they are passed to the output layer of the architecture. ACs driving policies predict the control actions at the output layer based on the 3-dimensional input images at each time step.

	Since our six selected DRL models work on a discrete as well as a continuous action space, the output layer of the architecture predicts nine distinct action values for the discrete action space and two float values for the continuous action space policies. These output actions can be summed into three vehicle control commands: Steer, Throttle, and Brake. Reverse is disabled for our experiments.
	
	%\begin{figure}[!htbp]
	%	\centering
	%	
	%	\includegraphics[width = 0.48\textwidth,height=0.075\textheight]{Action.pdf}
	%	\caption{Illustration of the action space with respect to discrete and continuous action space DRL algorithms.}
	%	\label{fig:Action}
	%\end{figure} 

\subsection{Reward functions}\label{sec:Reward}
As described in Section~\ref{sec:Multiagent}, each agent in a POMDP setting is following MDP. Thus, the driving policies receive reward $R$ at each time step of the simulated environment while collecting trajectories of the tuple $(S, R, A)$. $R$ is the reward value returned while performing action $A$ on the current state observations $S$ which helps in improving the driving policies $\pi_{AC}$. 

One of the contributions of this paper is the design of a detailed reward function with multiple objectives for learning ACs in a closed-loop multi-agent urban driving environments. For our experiments, we define the reward function $R_{AC}$ used by the DRL AC policies $\pi_{AC}$ during the training phase. 

	$R_{AC}$ can be formulated as:
	\begin{align*}
	R_{AC} =
	\begin{cases}
	- 50.0(CV_t \land CO_t \land CP_t) & \textcolor{purple!75}{\text{\textit{Safety}}}\\
	+ 10(\Delta D + F_t)& \textcolor{purple!75}{\text{\textit{Efficiency}}}\\
	- 0.5(OS_t)  & \textcolor{purple!75}{\text{\textit{Lane Keeping}}}\\
	+ \phi & \textcolor{purple!75}{\text{\textit{Penalty Constant}}}\\
	\end{cases}       
	\end{align*}

	%\begin{align*} 
	%R_{AC} = (D_{t-1} - D_{t}) + (F_t)/10 -100.0(CV_t + CO_t) 
	%\\  - 0.5(IO_t)  + \beta
	%\end{align*}
 $R_{AC}$ aims to maximize the driving performance of an AC agent by keeping the following objectives in check:

\textbf{\textit{Safety:}} The first objective of $R_{AC}$ includes $CV_t$, $CO_t$, and $CP_t$ that represent a boolean value \{0,1\} for collision with another vehicle, road objects, and pedestrian respectively. To ensure safety, AC tries to avoid any type of collision based on past rewards.

\textbf{\textit{Efficiency:}} If driving efficiently, an AC agent gets positive rewards on minimizing the distance to goal state compared to the previous timestep i.e. $\Delta D = D_{t-1} - D_{t}$ and speed of the driving agent as as $F_t$.

\textbf{\textit{Lane Keeping:}} For ensuring lane keeping, $OS_t$ refers to the boolean value represented by a boolean value \{0,1\} for minimizing offroad steering errors of the AC model.

\textbf{\textit{Penalty Constant:}} $\phi$ at the end of the reward function is a penalty constant used to encourage AC models to explore the environment and making decisions while driving. This helps in situations where DRL agents get stuck and remain indecisive in making decisions, and therefore not able to move and learn. 
% in driving conditions in order to have a kick start in exploring the environments and making decisions. 

 \subsection{Hyperparameters}\label{sec:Hyperparameters}

There are some common hyperparameter configurations called \textit{model hyperparameters} used for defining the training and testing of the driving DRL policies as described in Section~\ref{sec:NN}. As for the \textit{algorithmic hyperparamters}, we first select hyperparameters by going through the literature on the available implementations for DRL in AD. Since we are using six different DRL implementations for training their driving policies, each of the algorithms requires separate hyperparameter tuning. Therefore we take advantage of a hybrid hyperparameter tuning algorithm known as Population Based Training (PBT)~\cite{PBT} during the training of each DRL AC agent. 
%PBT helps in going through all possible hyperparameter options for having the best driving behavior during episodes. 
Each DRL is started with an initial configuration set of hyperparameters and with the help of PBT each DRL agent explores the possible configurations, leading to different training and episodic runs. The details of the hyperparameters for all the AC agents are provided in the GitHub repository~\footnote{https://github.com/AizazSharif/Benchmarking-QRS-2022}.

The extensive details of the key hyperparameters used in the training phase of the ACs are shown in Table~\ref{tab:Hyperparameters}. Based on a DRL algorithm and its interaction with the environment, the number of episodes varies between 650 and 800 episodes, and steps per training iteration also vary accordingly as shown in the Table. 
%On the other hand, during the testing phase, we run 50 total episodes, each having 5000 simulation steps in all environment settings per driving agent as explained in Section~\ref{sec:Results}.
	%\begin{table}[htbp]
	%	\caption{Hyperparameters for the training of AC the adversarial driving agents. } \label{tab:Hyperparameters} 
	%	
	%	\begin{center}
	%		\resizebox{!}{1.0cm}{
	%			\begin{tabular}{llllllll}
	%				
	%				\toprule
	%				
	%				\textbf{Hyperparameter} & \textbf{PPO} & \textbf{A2C} & \textbf{A3C} & \textbf{IMPALA} & \textbf{DQN} & \textbf{DDPG} & \textbf{TD3} \\ \midrule
	%				Total Training Iterations & 200 & 100  \\ 
	%				Training Steps per episode & 2048 & 2048  \\ 				
	%				Total Training steps & 40000000 & 20000000 \\
	%				Learning Rate & 0.0005 & 0.0005   \\ 
	%				Batch Size & 128 & 128  \\ 
	%				Optimizer	& Adam & Adam   \\ \bottomrule
	%				
	%			\end{tabular}
	%		}
	%	\end{center}
	%\end{table}
	%\subsection{Driving scenario}\label{sec:Training}
	\begin{table}[!htbp]
		\caption{Hyperparameters for the training of AC driving agents. } \label{tab:Hyperparameters} 
		\centering % used for centering table
		\resizebox{\columnwidth}{!}{%
			\begin{tabular}
				{ p{3cm} p{3cm} p{3cm} }%{c c c } % centered columns (4 columns)
				\hline %inserts double horizontal lines
				
				AC Policy & Hyperparameters & value \\  % inserts table
				
				%heading
				\hline % inserts single horizontal line
				& Total Training Episodes            & 650  \\
				& Epochs per minibatch     & 8 \\ 
				& Value Loss Coefficient     & 1.0 \\ 
				& Entropy Coefficient     & 0.01 \\ 
				PPO & Total Training steps         & 1331200 \\
				& Learning Rate     & 0.001 \\
				& Clipping     & 0.3 \\ 
				& KL Initialization     & 0.3 \\ 
				& Lambda     & 0.99 \\
				& KL Target           & 0.03 \\ 
				
				\hline % inserts single horizontal line
				& Total Training Episodes            & 600  \\
				& Value Loss Coefficient     & 1.0 \\ 
				& Entropy Coefficient     & 0.01\\ 
				A3C & Total Training steps         & 1228800 \\
				& Learning Rate     & 0.005 \\
				& Lambda     & 1.0 \\
				& Gradient Clipping & 30.0 \\
				& Sample Async  & True \\  
				
				\hline % inserts single horizontal line
				& Total Training Episodes            & 800  \\
				& Value Loss Coefficient     & 0.5 \\ 
				& Entropy Coefficient     & 0.01\\ 
				IMPALA & Vtrace & True \\
				& Vtrace Threshold & 1.0 \\
				& Total Training steps   & 1638400 \\
				& Learning Rate     & 0.0001 \\
				& Gradient Clipping & 30.0 \\ 
				
				\hline % inserts single horizontal line
				& Total Training Episodes            & 800  \\
				& No. of Atoms   & 1 \\ 
				& Noisy Network   & False \\ 
				& Dueling DQN & True \\ 
				& N-step Q-learning   & 1 \\ 
				DQN & Initial $\epsilon$   & 1.0 \\ 
				& Final $\epsilon$   & 0.02 \\ 
				& Replay Buffer Size   & 400 \\ 
				& Entropy Coefficient     & 0.01\\ 
				& Total Training steps   & 1638400 \\
				& Learning Rate     & 0.0001 \\
				& Gradient Clipping & 40.0 \\  
				
				\hline % inserts single horizontal line
				& Total Training Episodes            & 800  \\
				& Policy Delay   & 1 \\ 
				& Target Noise   & 0.2 \\ 
				& Sigma ($\Sigma$) & 0.2 \\ 
				& Tau ($\tau$) & 0.002 \\ 
				& N-step Q-learning   & 1 \\ 
				DDPG & Initial Exploration Scale   & 1.0 \\ 
				& Replay Buffer Size   & 500 \\
				& Critics Learning Rate   & 0.001 \\
				& Actor Learning Rate   & 0.001 \\
				& Entropy Coefficient     & 0.01\\ 
				& Total Training steps   & 1638400 \\
				& Learning Rate     & 0.0001 \\ 
				
				\hline % inserts single horizontal line 
				& Total Training Episodes            & 790  \\
				& Policy Delay   & 1 \\ 
				& Target Noise   & 0.2 \\ 
				& Tau ($\tau$) & 0.0002 \\ 
				& Sigma ($\Sigma$) & 0.2 \\ 
				& N-step Q-learning   & 1 \\ 
				& Initial Exploration Scale   & 1.0 \\ 
				TD3 & Replay Buffer Size   & 500 \\
				& Critics Learning Rate   & 0.001 \\
				& Actor Learning Rate   & 0.001 \\
				& Entropy Coefficient     & 0.01\\ 
				& Total Training steps   & 1617920 \\
				& Learning Rate     & 0.0001 \\
				& Lambda     & 0.005 \\
				\hline % inserts single horizontal line		
				
				&Batch Size & 128 \\ 
				&Batch Mode & Complete Episodes \\ 
				& Epsilon ($\epsilon$) & 0.1 \\
				Common Parameters & Training Steps per episode     & 2048 \\                                                                               
				& Optimizer & Adam  \\                                                                             & 	 Discount Factor ($\gamma$)    & 0.99   \\ 
				\hline %inserts single line
			\end{tabular}
		}
		\label{table:Hyperparameters} % is used to refer this table in the text
	\end{table}
	
	%\section{Experimental Evaluation} \label{sec:Experimental}
	\section{Comparative study} \label{sec:Experimental}
In this section, we present a comparative study of the performance of six DRL algorithms for AD, including the description of the simulation frameworks for the training and testing of AC agents.  

The research questions in our work evaluate:

\textbf{RQ1:} Which DRL-based ACs act better (or worse) in a single-agent as well as a multi-agent competitive scenario when trained in a multi-agent-only scenario?

\textbf{RQ2:} Which DRL-based ACs are more successful in fulfilling multiple driving objectives throughout different environments?

In RQ1, we evaluate the driving performance of victim ACs using six \textbf{Driving Performance Metrics}:
\begin{itemize}
	\item \textit{CV}: the amount of collision with another driving vehicle
	\item \textit{CO}: the amount of collision with road objects
	\item \textit{CP}: the amount of collision with pedestrians
	\item \textit{OS}: offroad steering percentage from its driving lane
	\item \textit{TTFC}: the time it takes to have the first collision
	\item \textit{SPEED}: the forward speed of the AC agent
%	\item \textit{DISTANCE}: the distance covered by an AC agent.
\end{itemize}

For $CV$, $CO$, $CP$, and $OS$ we calculate the percentage of error within each episode (as a value between 0 and 1). In each testing episode per environment, we run 5000 simulation steps and at the end of 50 episodic test runs, we compute the average error rate for each metric across all the episodes. $TTFC$ on the other hand shows the time in seconds it takes to detect the first collision in a testing episode. $SPEED$ is visualized to plot the average forward speed (in km/h) by each DRL-based ACs. 

 Whereas in RQ2, we use the same Driving Performance Metrics to measure the success rate of DRL ACs with respect to \textbf{\textit{Safety}}, \textbf{\textit{Efficiency}}, and \textit{\textbf{Lane Keeping}} across all environments.
\begin{itemize}
	\item \textbf{\textit{Safety}} is calculated as an average among all collision based metrics such as $CV$, $CO$, and $CP$.
	\item \textbf{\textit{Efficiency}} is computed by using $SPEED$ and a seventh metric $DISTANCE$.
	\item  \textit{\textbf{Lane Keeping}} is measured as an average among offroad steering metric values taken as $OS$.
\end{itemize}

For calculating $DISTANCE$, we define a goal state for each driving condition covered by each DRL policy during testing. This is required since we can only rely on current episodic results during testing.
$DISTANCE$ can be formulated as:

\begin{align*}
DISTANCE_t = \sqrt { \left( x_{goal}-x_{spawn}\right)^2 + \left( y_{goal}-y_{spawn}\right)^2}  -  \\ \sqrt { \left( x_{goal}-x_{current}\right)^2 + \left( y_{goal}-y_{current}\right)^2}
\end{align*}

where it measures the difference between euclidean distances from the goal to the spawn state versus the ending states of the episode.

We analyze which DRL-based AC agent achieves each learning objective as described in Section~\ref{sec:Reward}.

\subsection{Driving Environment}\label{sec:env}
\begin{figure}[!htbp]
	\centering
	
	\includegraphics[width = 0.49\textwidth,height=0.23\textheight]{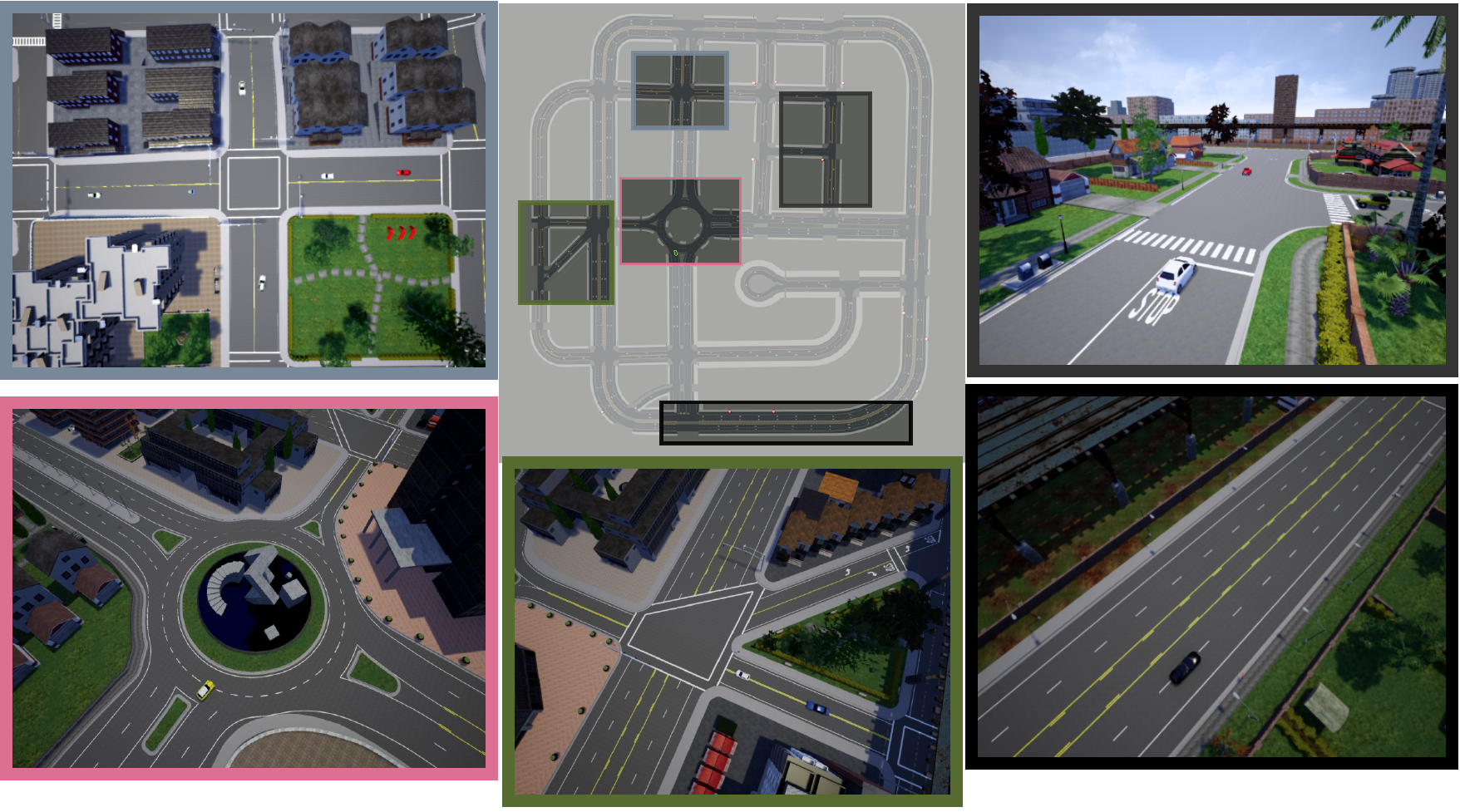}
	\caption{Illustration of Town03 Carla urban driving environment. 
		Bottom middle highlighted area on the map represents \textit{env\_1} (Straight), top right highlighted area on the map represents \textit{env\_2} (Three Way intersection) and top middle highlighted area displays \textit{env\_3} (Four-way intersection).  Middle highlighted area on the same map shows \textit{env\_4} (Roundabout) while bottom left highlighted area represents \textit{env\_5} (Merge). }
	\label{fig:Step2}
\end{figure}

To train our AC agents in a partially observable urban environment we use \textit{Town03} map from Carla with Python API~\cite{town}. This environment is configured for training multi-agent ACs and testing the same ACs in a single and multi-agent scenario. 
%The same environment is also used for training the adversaries against the best-performing AC. 
The details of the training and testing configurations for our experiments are mentioned in Section~\ref{sec:Steps}. We first provide brief descriptions of the driving environments displayed in the Figure~\ref{fig:Step2}.

\subsubsection{\textbf{Driving Environment 1 (\textit{env\_1}})}\label{sec:env1}
We use \textit{Straight} road from the Town03 map. The driving setting is simple yet suitable especially for validating multi-agent AC policies in order to see their lane-keeping capability.

\subsubsection{\textbf{Driving Environment 2 (\textit{env\_2}})}\label{sec:env2}
We use \textit{Three Way intersection} (also known as T-intersection) from the Town03 map. The driving setting is perfect for validating multi-agent AC policies in a scenario where multiple driving cars, as well as pedestrians, are faced. 
%All the independent non-communicating driving agents are spawned closer to the intersection in all three scenarios as mentioned in Section~\ref{sec:Steps}.

\subsubsection{\textbf{Driving Environment 3 (\textit{env\_3}})}\label{sec:env3}

The environment has independent non-communicating agents spawned close to the \textit{four-way intersection} throughout the testing scenarios. The choice of a four-way intersection as a driving scenario is based on its higher complexity for an AC agent. Similar to \textit{env\_2}, this driving environment also contains pedestrians. 	

\subsubsection{\textbf{Driving Environment 4 (\textit{env\_4}})}\label{sec:env4}

The environment has independent non-communicating agents spawned close to the \textit{Roundabout} throughout the training and testing scenarios. The choice of the roundabout as a driving scenario is based on its higher complexity for an AC agent. 

\subsubsection{\textbf{Driving Environment 5 (\textit{env\_5}})}\label{sec:env5}

The environment is called \textit{Merge} where two roads merge to a single point. The choice of merge as a driving scenario is also based on its higher complexity for an AC agent. 

	%\subsection{Experimental Setup}\label{sec:metrics}
	%For evaluating the performance of each DRL AC policy $\pi_{AC}$, we look at four metrics: i) the amount of collision with another driving vehicle, ii) the amount of collision with road objects, iii) the offroad steering percentage from its driving lane, and iv) the forward speed of the AC agent. The first 3 metrics are used together for showing the overall driving safety capabilities of the AC policy, while the fourth metric is used for understanding in-depth the driving behavior of specific AC agents.
	
	\subsection{Experimental Setup}\label{sec:Steps}
	
	\subsubsection{\textbf{Scenario 1: Training and testing of multi-agent ACs}}\label{sec:Step1}
\begin{figure}[htbp]
		\captionsetup[subfigure]{labelformat=empty}
		\centering     %%% not \center
		\subfloat[PPO]{\includegraphics[width=43.5mm]{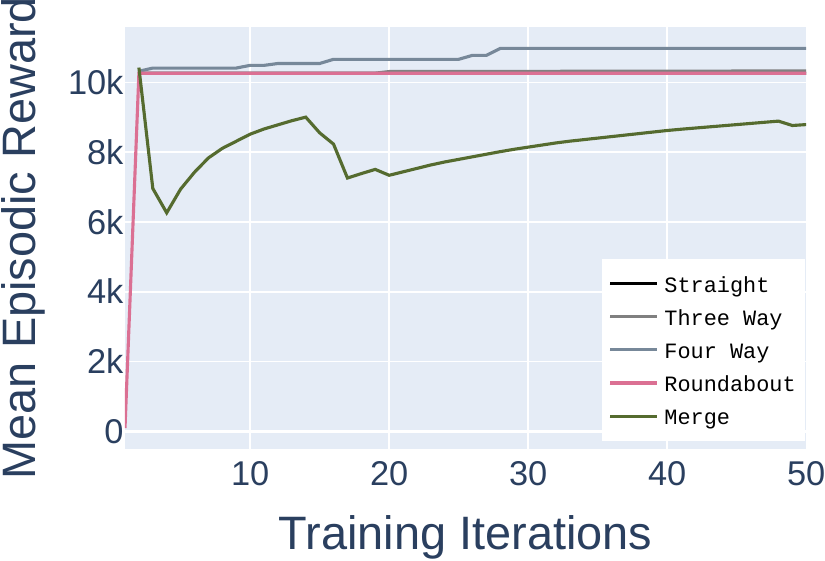}}
		\subfloat[A3C]{\includegraphics[width=42mm]{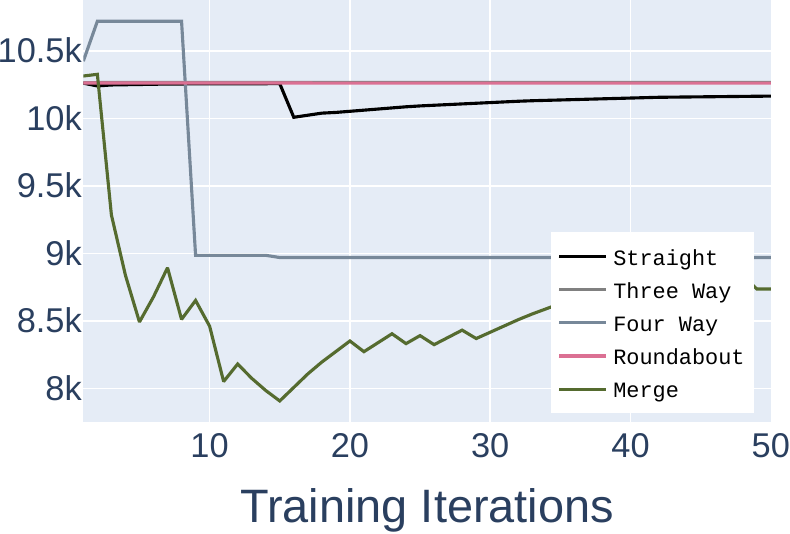}}\\
		\subfloat[IMPALA]{\includegraphics[width=44.5mm]{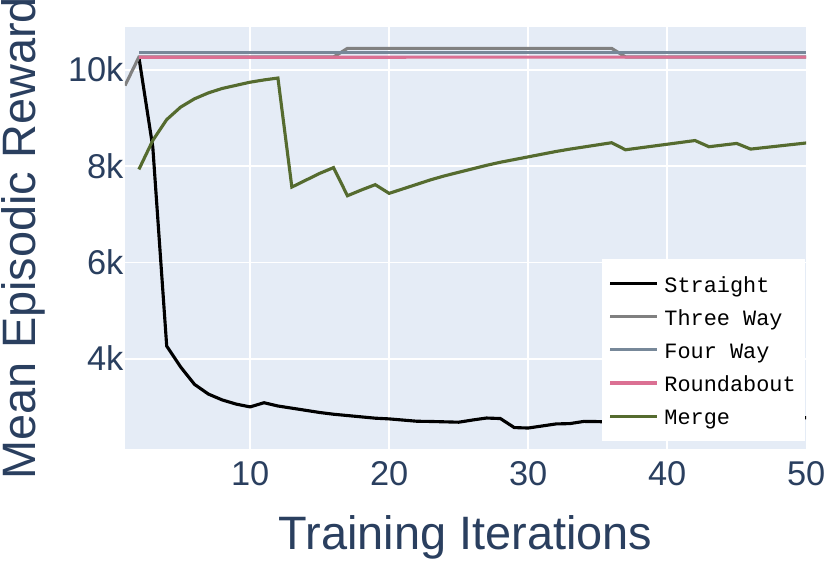}}
		\subfloat[DQN]{\includegraphics[width=41.5mm]{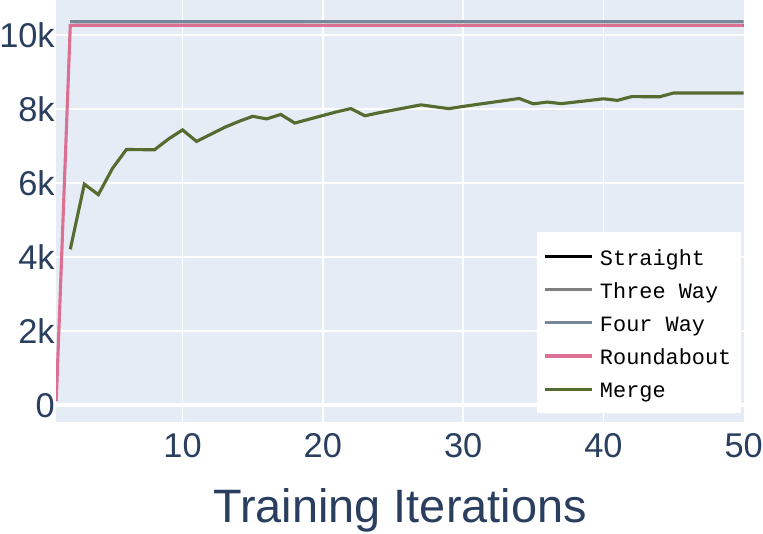}}\\
		\subfloat[DDPG]{\includegraphics[width=43.5mm]{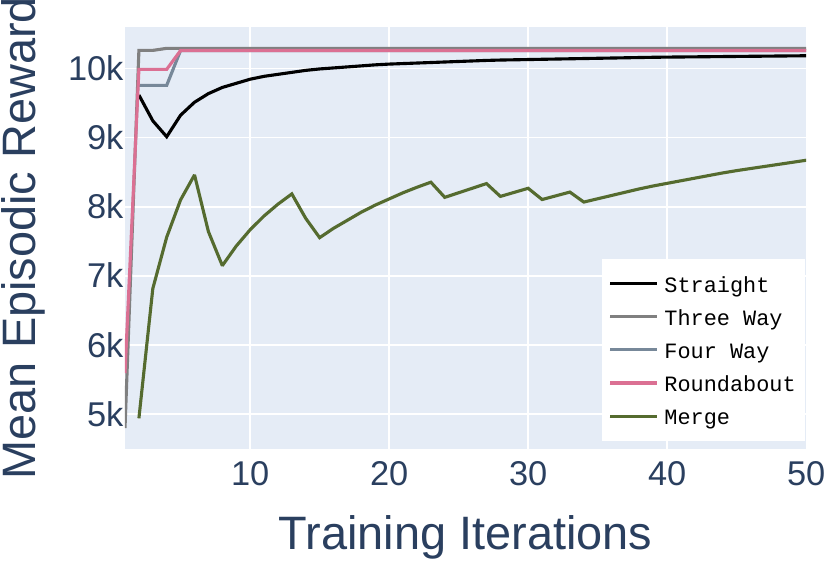}}
		\subfloat[TD3]{\includegraphics[width=41mm]{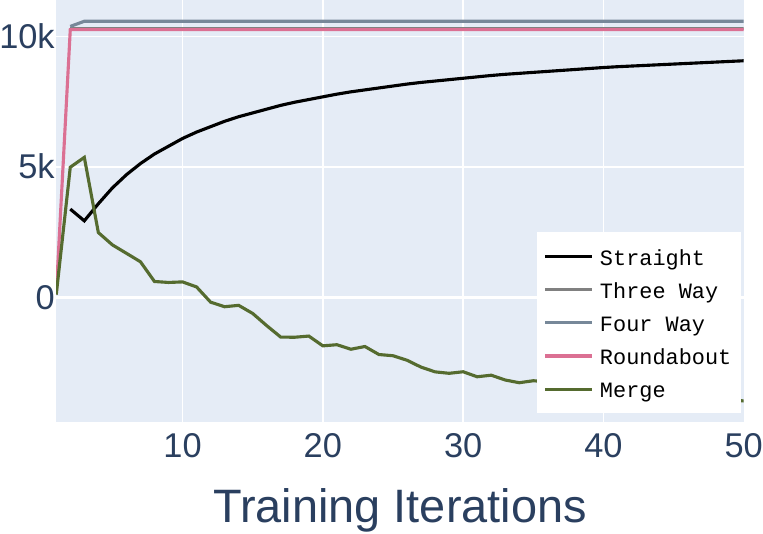}}
		\caption{\label{fig:Step1}Convergence of DRL AC policies in \textit{Scenario 1}. Each plot represents DRL algorithm trained in all 5 driving environments. The first four plots shows mean episodic reward for PPO, A3C, IMPALA, and DQN from a discrete action space. The last two plots illustrate mean episodic
reward for DDPG and TD3 algorithms from a continuous action space.}
	\end{figure}

	First, we train AC policies $\pi_{AC}$ in a multi-agent driving environment. There are DRL-based, as well as a few auto-controlled cars driving around each of the AC agents representing human drivers from real-life scenarios. We train $\pi_{AC}$ policies using the reward function $R_{AC}$ described in Section~\ref{sec:Reward}. The performance of AC policies that are trained in the first scenario is shown in Figure~\ref{fig:Step1}. 
	
	%By training the $\pi_{AC}$ policies for different number of iterations, almost every AC policy converges faster except the $\pi_{IMPALA}$ and $\pi_{TD3}$ based policy. $\pi_{A3C}$ based DRL policies get to a state stable max episodic reward after a few training episodes. 

	%By training all 7 DRL-based ACs in~\ref{sec:Step1}, we bring them in a multi-agent scenario where they drive next to each other, as well as around auto control drivers. Unlike single-agent driving, every AC acts as an independent non-communicating competitive driving agent, as illustrated in Figure~\ref{fig:Step2_4}\textbf{(a)}. 
	
	Next, for testing and validating ACs in a multi-agent \textit{Scenario 1} all six DRL-based ACs drive next to each other, as well as around auto-controlled cars, therefore acting as independent non-communicating competitive driving agents, illustrated in Figure~\ref{fig:Step2_4}\textbf{ (a)}. The driving performance of each AC is evaluated against the six Driving Performance Metrics.
	
	\subsubsection{\textbf{Scenario 2: Testing of single-agent ACs}}\label{sec:Step2}
	
\begin{figure}[!htbp]
	\centering
	
	\includegraphics[width = 0.44\textwidth,height=0.10\textheight]{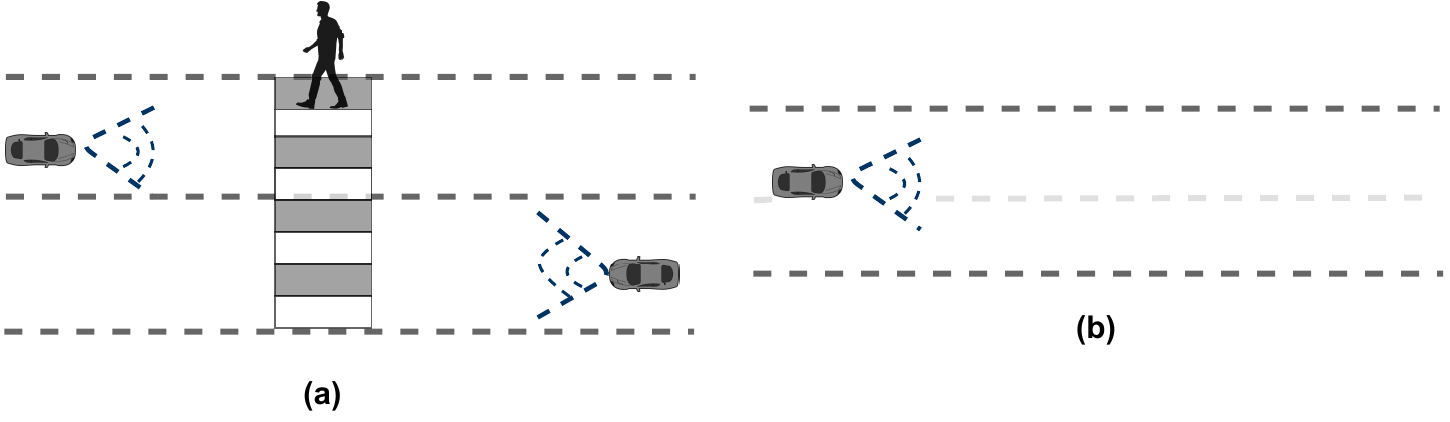}
	\caption{Illustration of two different scenarios for the testing phase. The left figure (a) shows a competitive multi-agent testing environment for all of the five driving environments. \textit{env\_2} and \textit{env\_3} additionally consists of pedestrians as well. Right figure (b) displays a single-agent driving scenario for testing every DRL-based AC agent individually.}
	\label{fig:Step2_4}
\end{figure}

	After training all six DRL-based ACs for a number of episodes in a multi-agent \textit{Scenario 1}, we validate the driving performance of each AC policy in a single-agent driving environment, illustrated in Figure~\ref{fig:Step2_4}\textbf{ (b)}. Testing in a single-agent scenario is important in order to validate the driving capability of AC policies in similar driving situations but with no cars around. Just like in \textit{Scenario 1}, ACs' driving performance is validated based on the six Driving Performance Metrics.

 Both scenarios are illustrated in Figure~\ref{fig:Step2_4}.

\subsection{Simulation Setup}\label{sec:Simulation}
The proposed benchmarking framework uses the following open libraries/frameworks:

\textit{CARLA}~\cite{R51} is an urban driving simulation framework designed for training and validating autonomous vehicles. CARLA is famous for its highly integrated Python API and access to high-fidelity urban driving environments. We use the 0.9.4 version.

\textit{RLlib}~\cite{R54} is a very fine-tuned and scalable DRL framework library. RLlib gives the opportunity to access more than one DRL policy graph and its hyperparameters for creating a non-shared multi-agent system. We use versions 0.8.0 for IMPALA and 0.8.5 for the rest of the 5 DRL algorithms.

\textit{Macad-gym}~\cite{R52} is an open-source framework that connects CARLA, RLlib, and Open AI's Gym toolkit~\cite{gym}. We have modified the framework by adding competitive multi-agent driving functionalities required for our experiments. We use the 0.1.4 version.

\textit{Tensorflow}~\cite{R53} is one of the leading frameworks used to create deep learning-based algorithms. We use version 2.2.0 within the RLlib library.

	\section{Results \& Analysis}\label{sec:Results}

	In this section, we discuss the experimental results for testing DRL AC policies. For collecting the results, we run 50 testing episodes in both single and multi-agent scenarios to take an average among the Driving Performance Metrics explained in Section~\ref{sec:Experimental}. In each episode, we use 5000 simulation steps in \textit{env\_1} to \textit{env\_5} in order to test the performance of every AC driving policy.

	\subsection{RQ1: \textbf{Testing AC policies in a single and multi-agent scenario}} \label{sec:RQ1}
	
	We first look into the performance comparison among the 4 discrete and 2 continuous action space DRL AC policies. Each of the AC agents is trained with a various number of episodes and their model convergence performance has been discussed in Section~\ref{sec:Step1}. Now we use the same training environment and analyze their driving behavior in both single and multi-agent scenarios.
	
\begin{figure*}[htbp]
		\captionsetup[subfigure]{labelformat=empty}
		
		\centering     %%% not \center
		\subfloat[PPO env\_1]{\includegraphics[width=27mm]{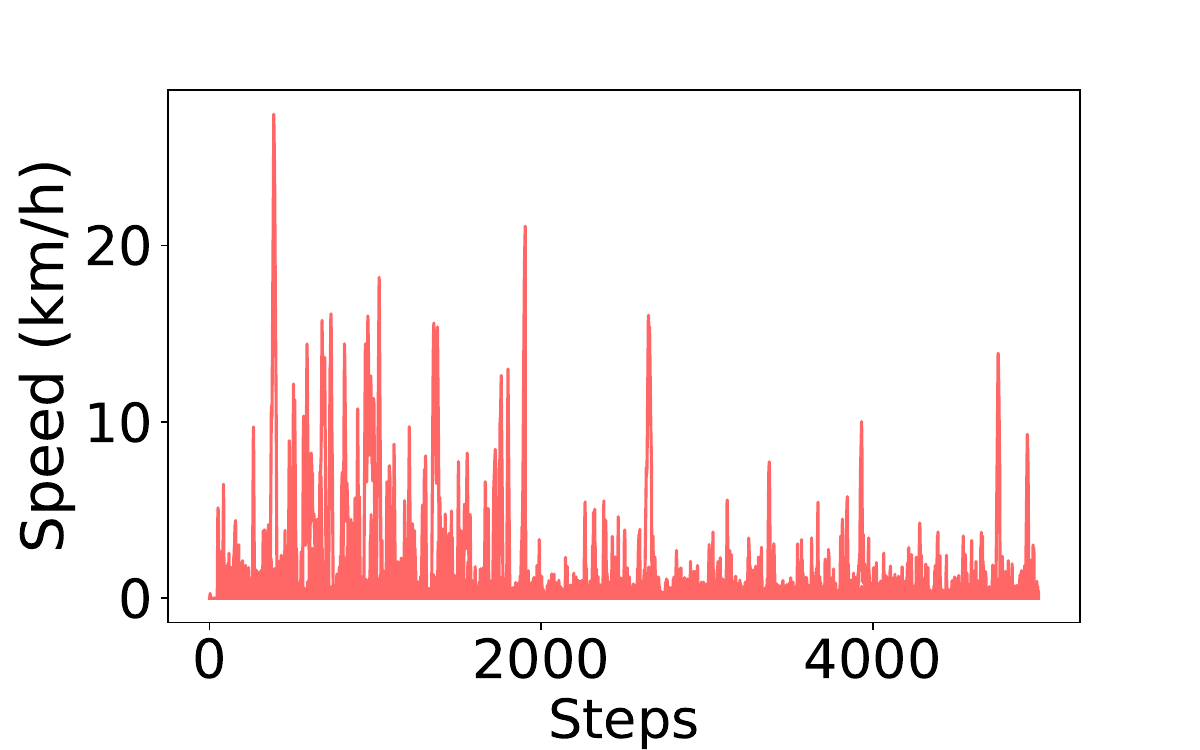}}
		\subfloat[A3C env\_1]{\includegraphics[width=27mm]{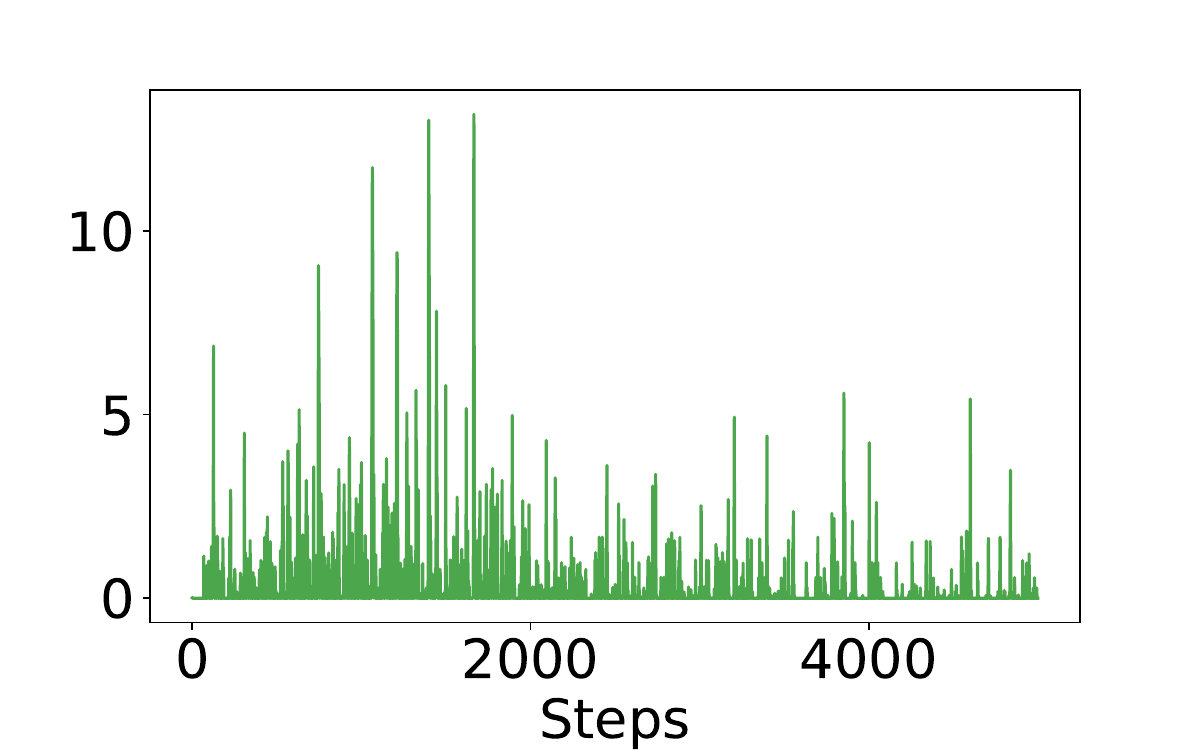}}
		\subfloat[IMPALA env\_1]{\includegraphics[width=27mm]{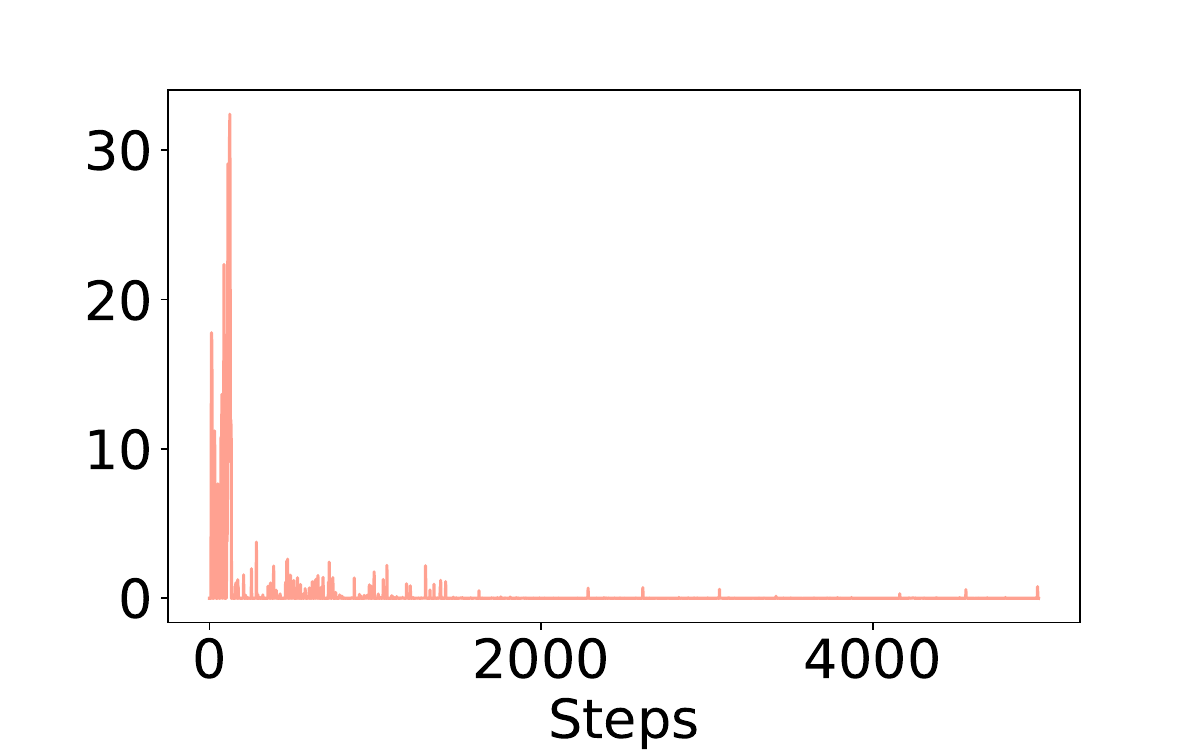}}
		\subfloat[DQN env\_1]{\includegraphics[width=27mm]{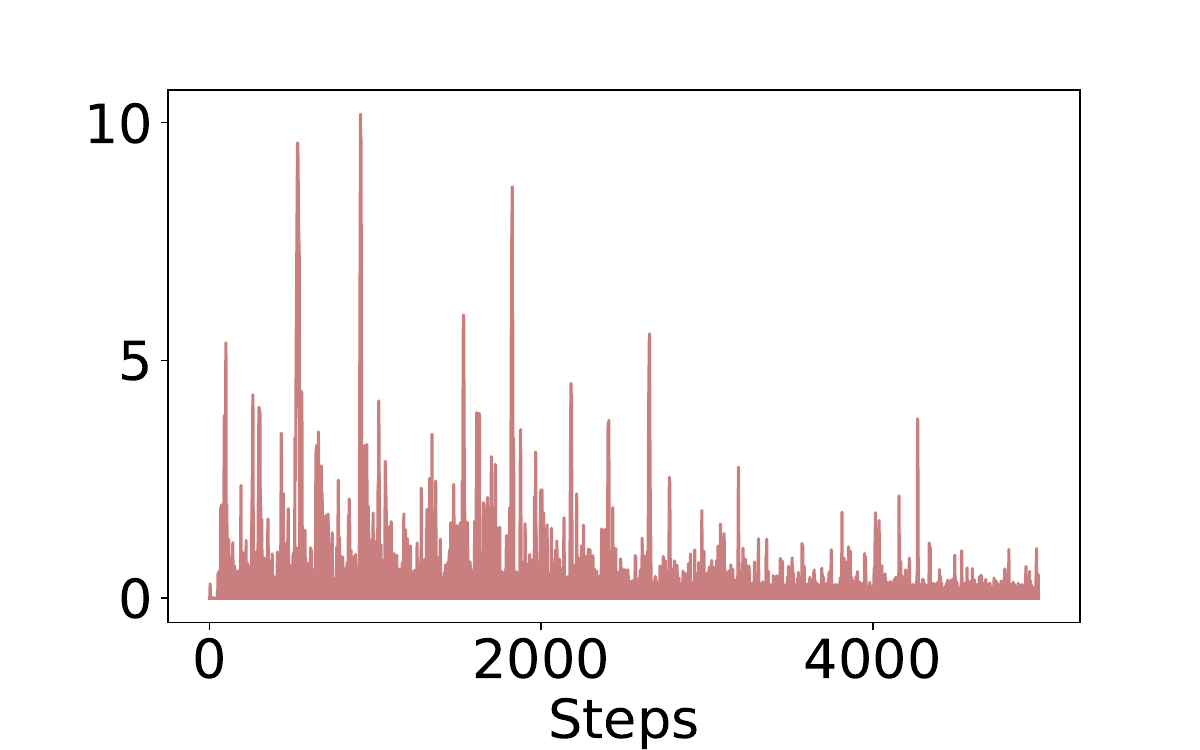}}
		\subfloat[DDPG env\_1]{\includegraphics[width=27mm]{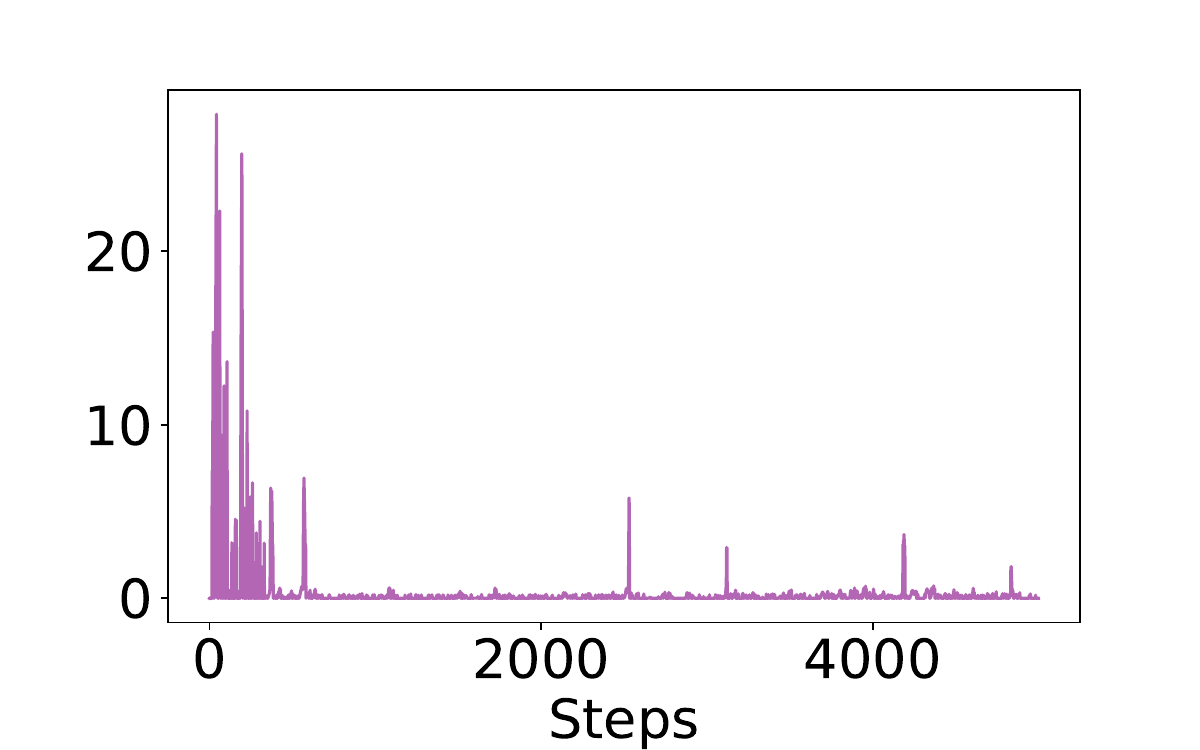}}
		\subfloat[TD3 env\_1]{\includegraphics[width=27mm]{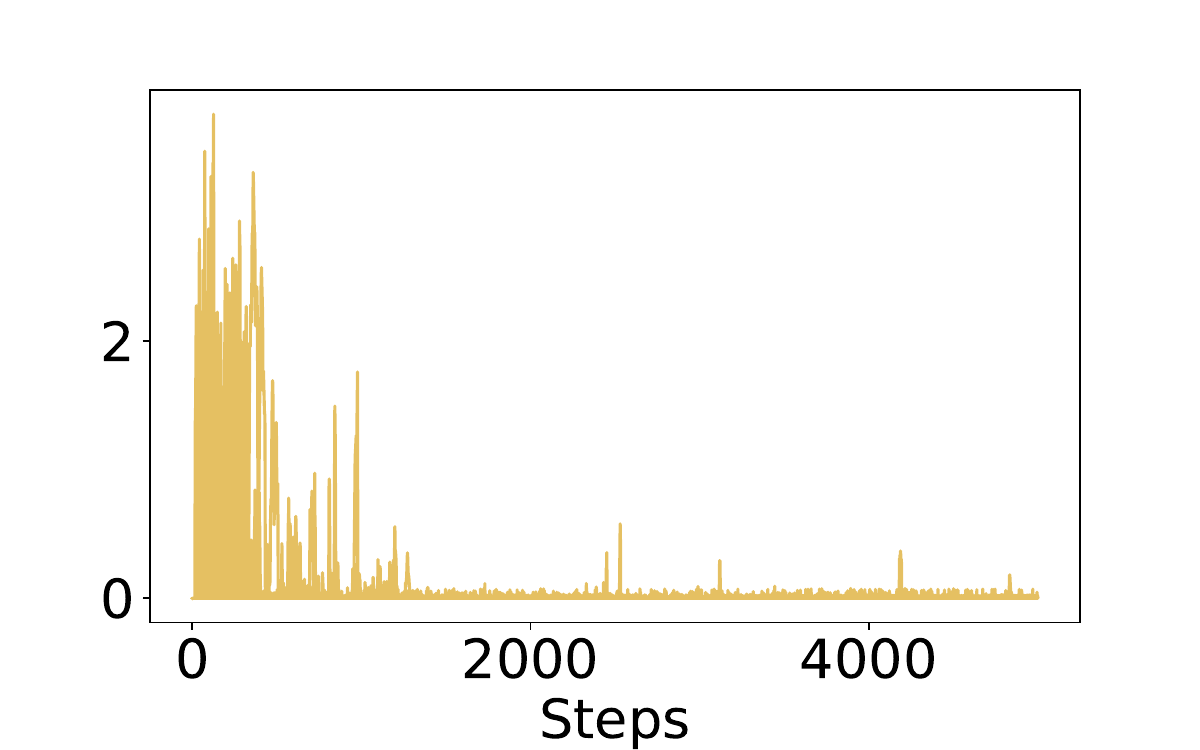}}
		
		\subfloat[PPO env\_2]{\includegraphics[width=27mm]{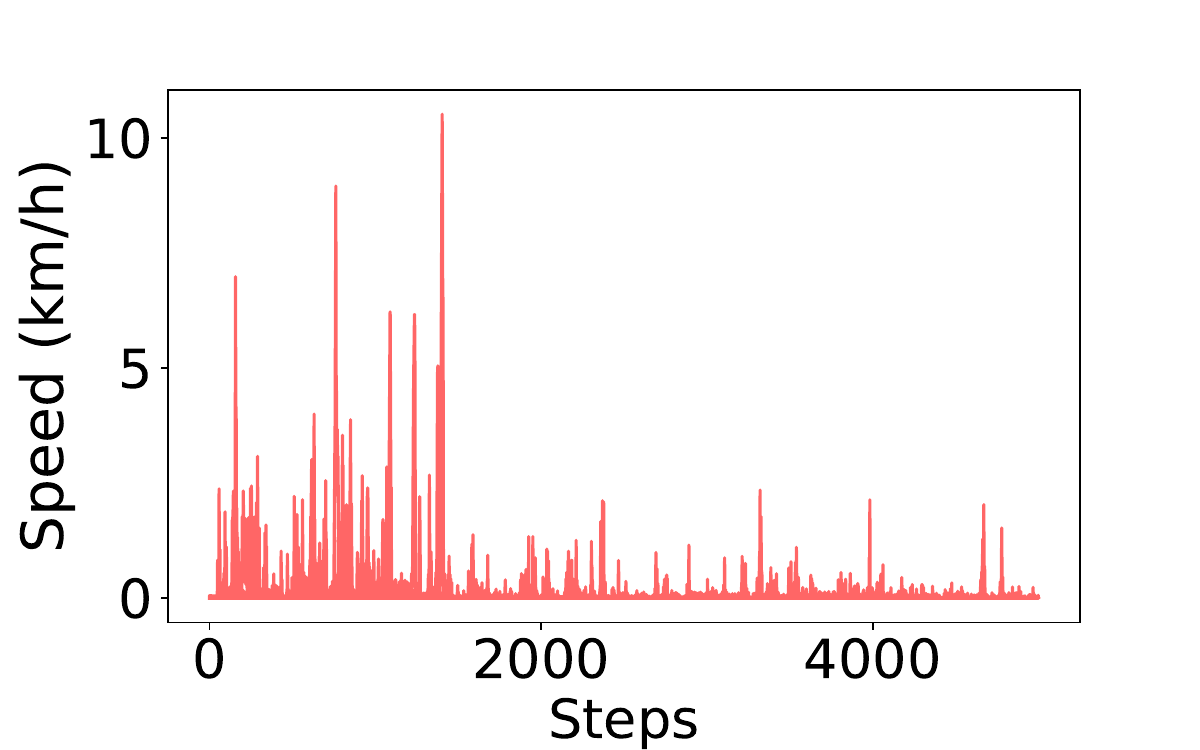}}
		\subfloat[A3C env\_2]{\includegraphics[width=27mm]{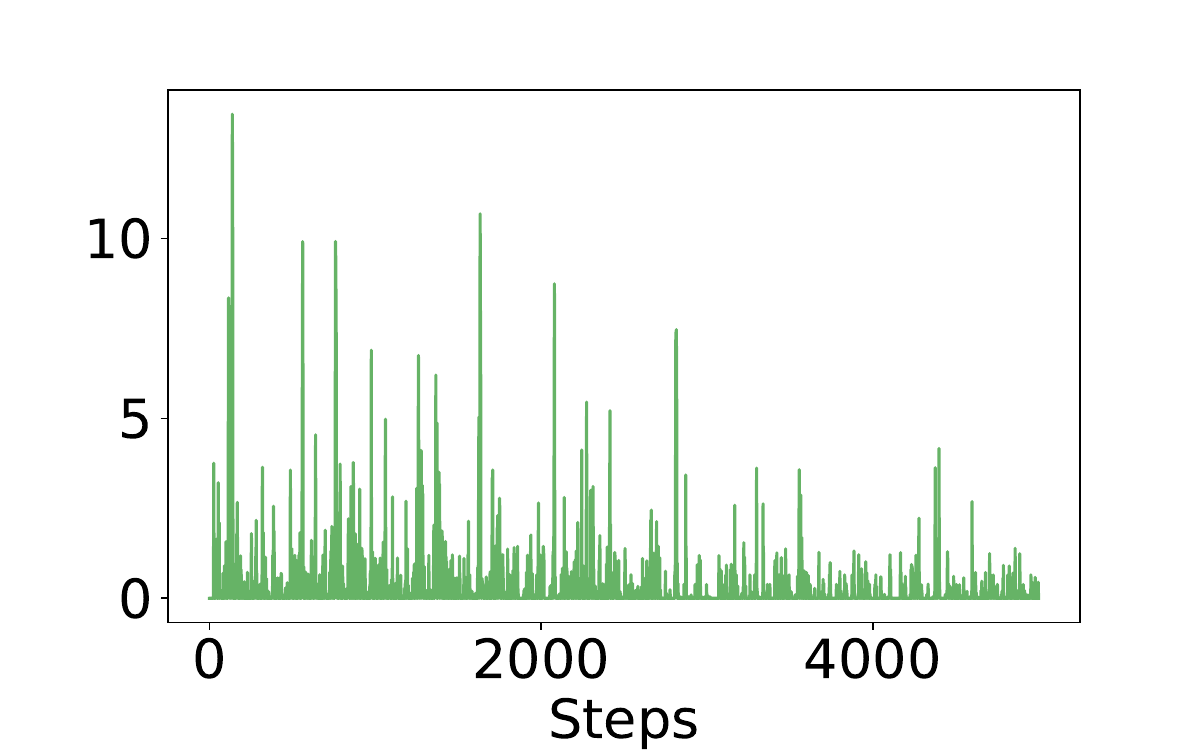}}
		\subfloat[IMPALA env\_2]{\includegraphics[width=27mm]{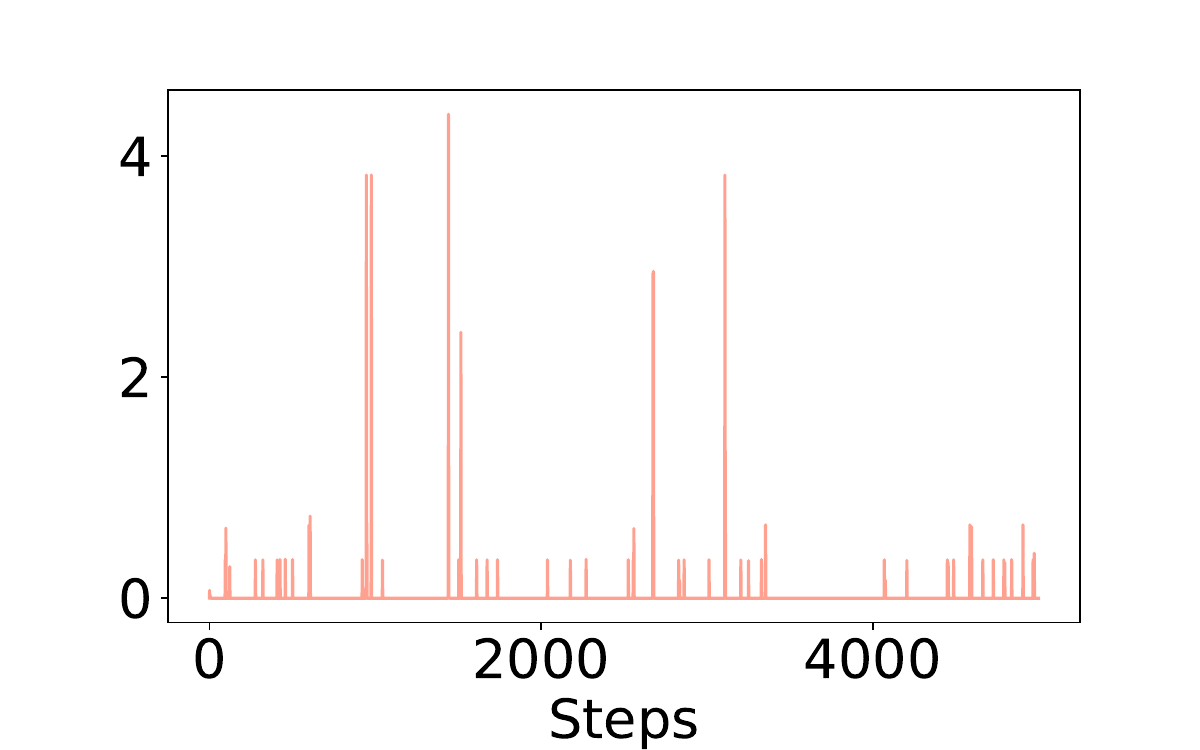}}
		\subfloat[DQN env\_2]{\includegraphics[width=27mm]{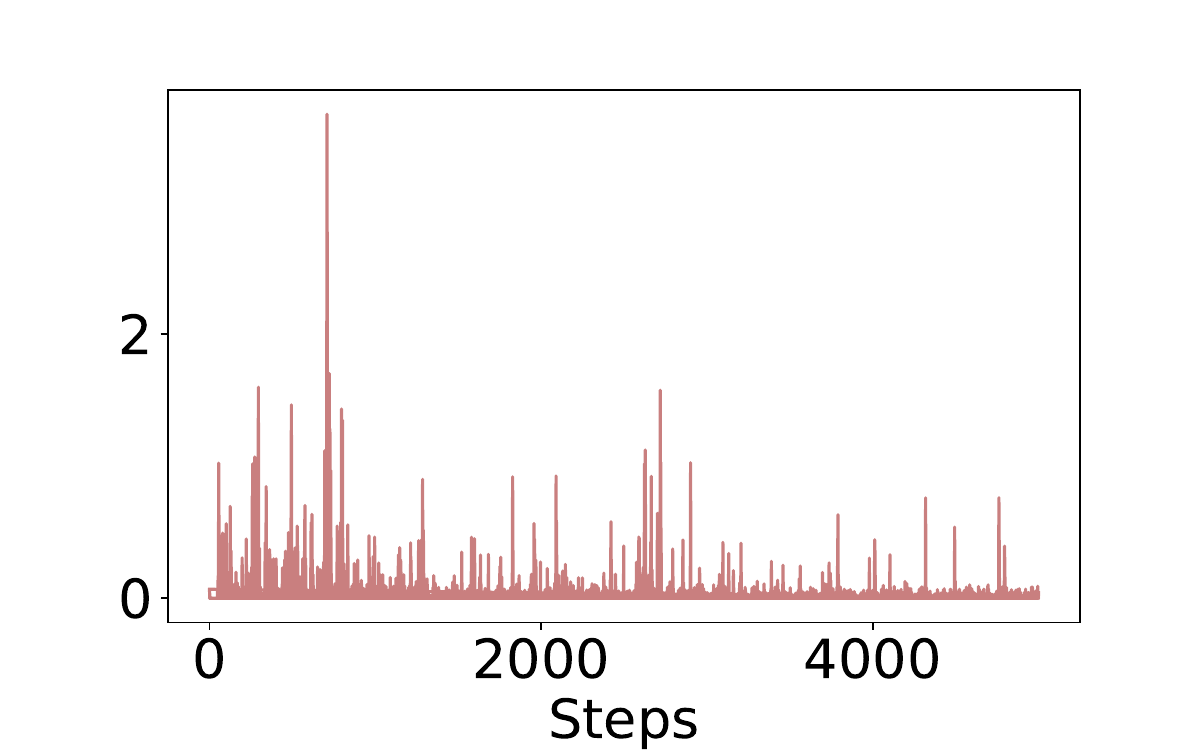}}
		\subfloat[DDPG env\_2]{\includegraphics[width=27mm]{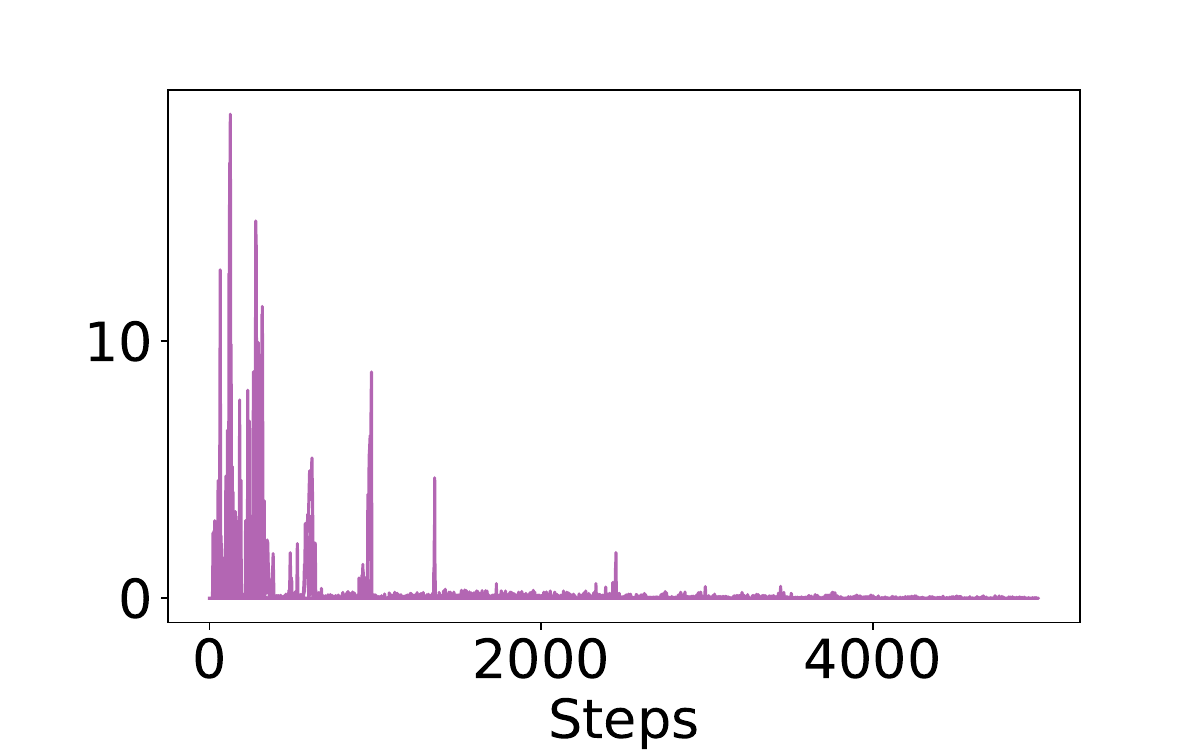}}
		\subfloat[TD3 env\_2]{\includegraphics[width=27mm]{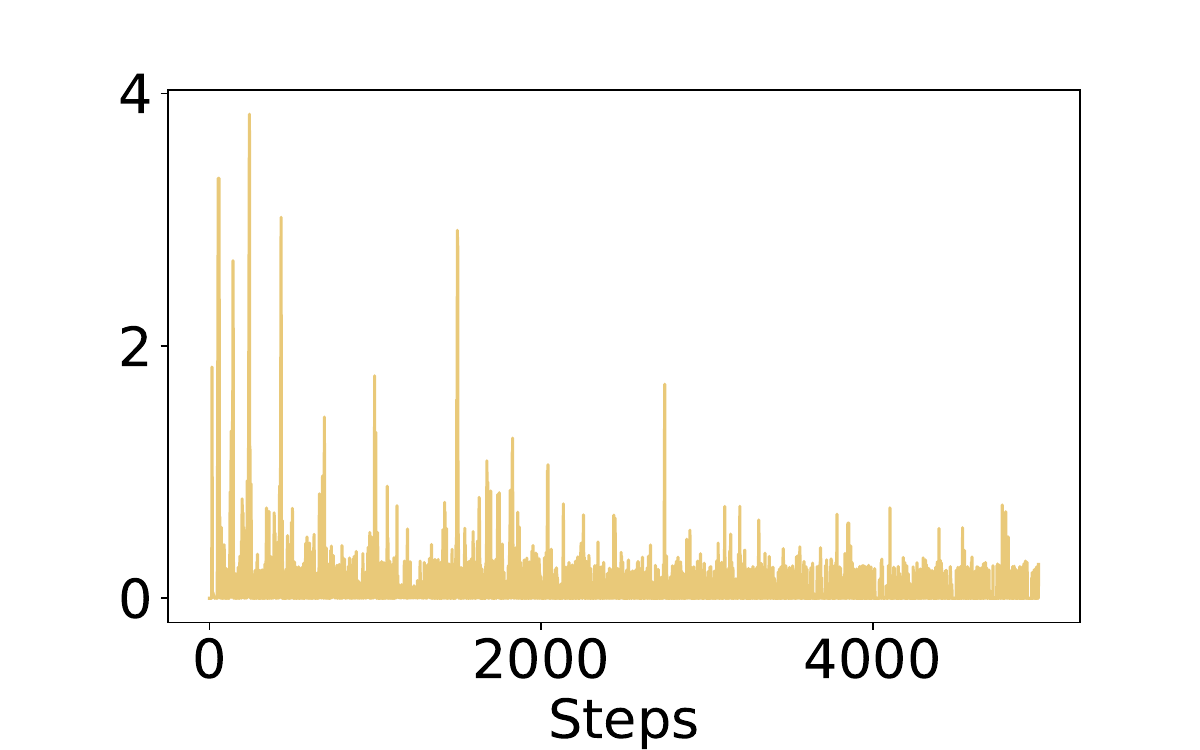}}
		
		\subfloat[PPO env\_3]{\includegraphics[width=27mm]{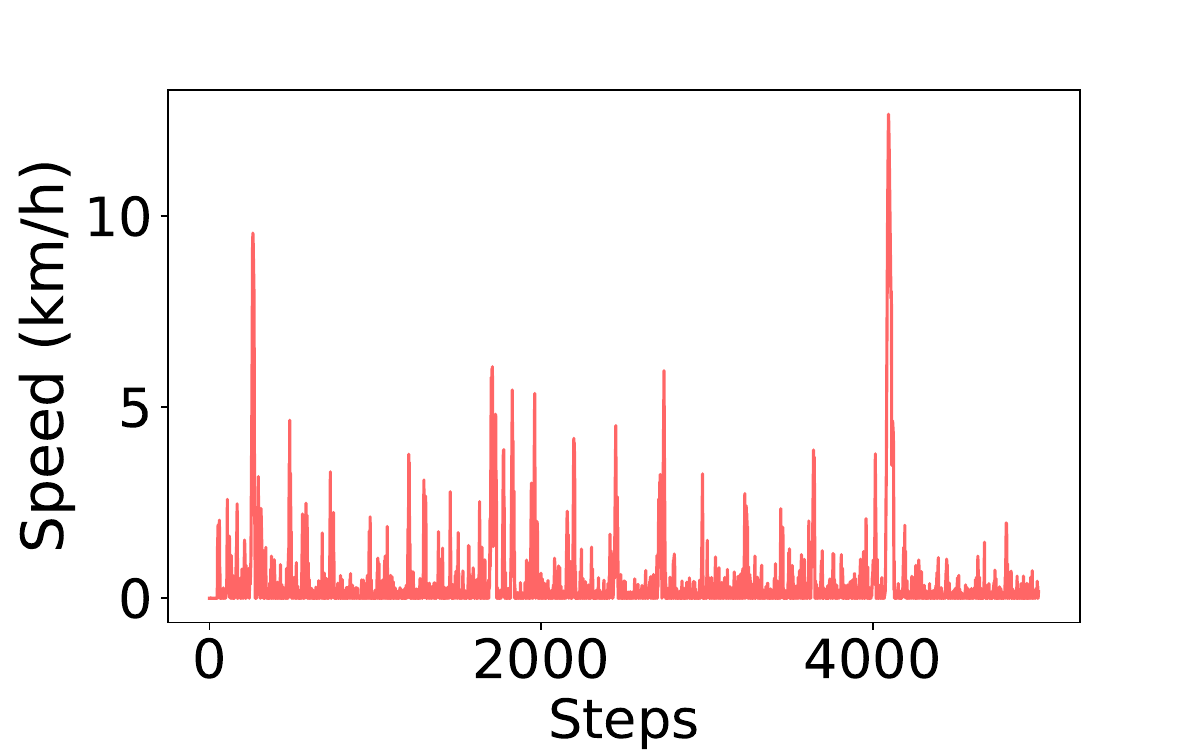}}
		\subfloat[A3C env\_3]{\includegraphics[width=27mm]{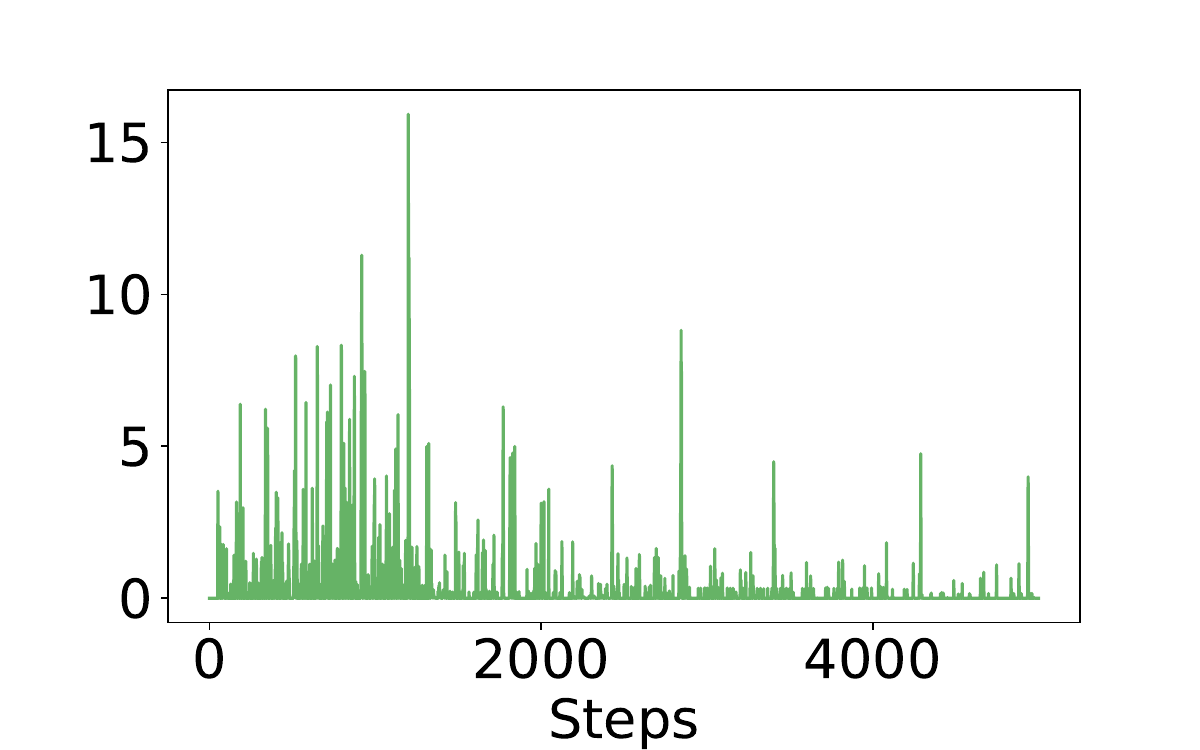}}
		\subfloat[IMPALA env\_3]{\includegraphics[width=27mm]{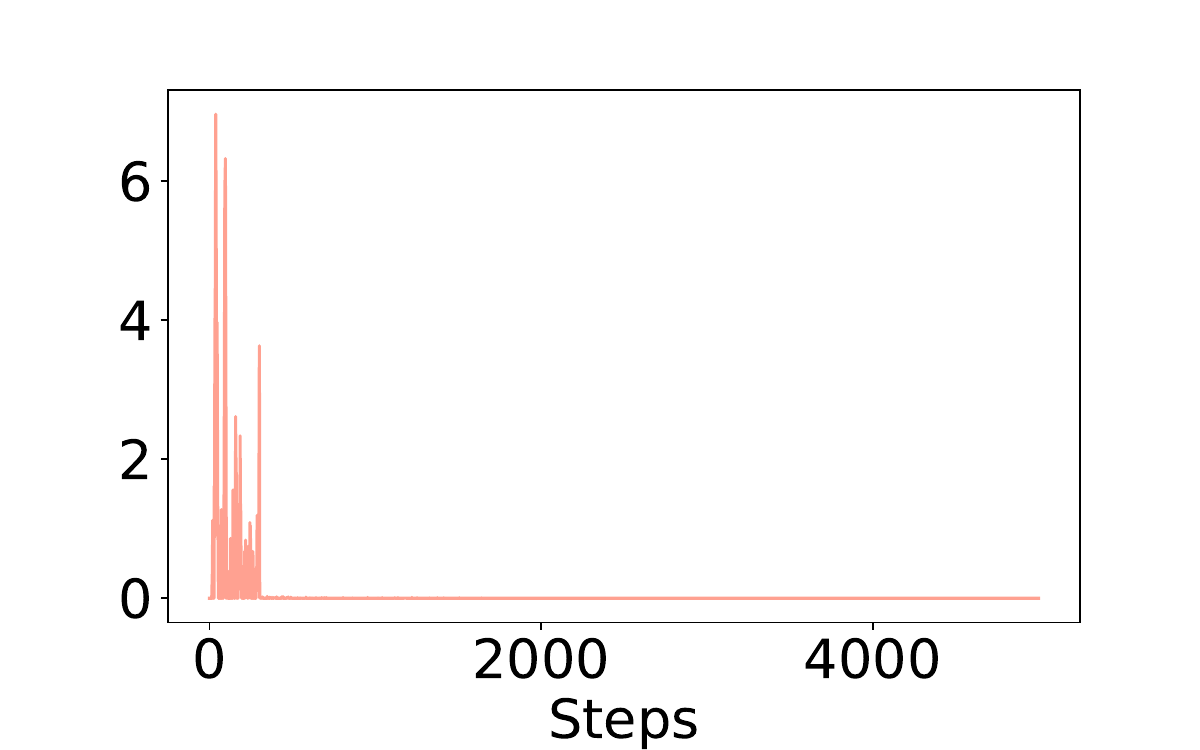}}
		\subfloat[DQN env\_3]{\includegraphics[width=27mm]{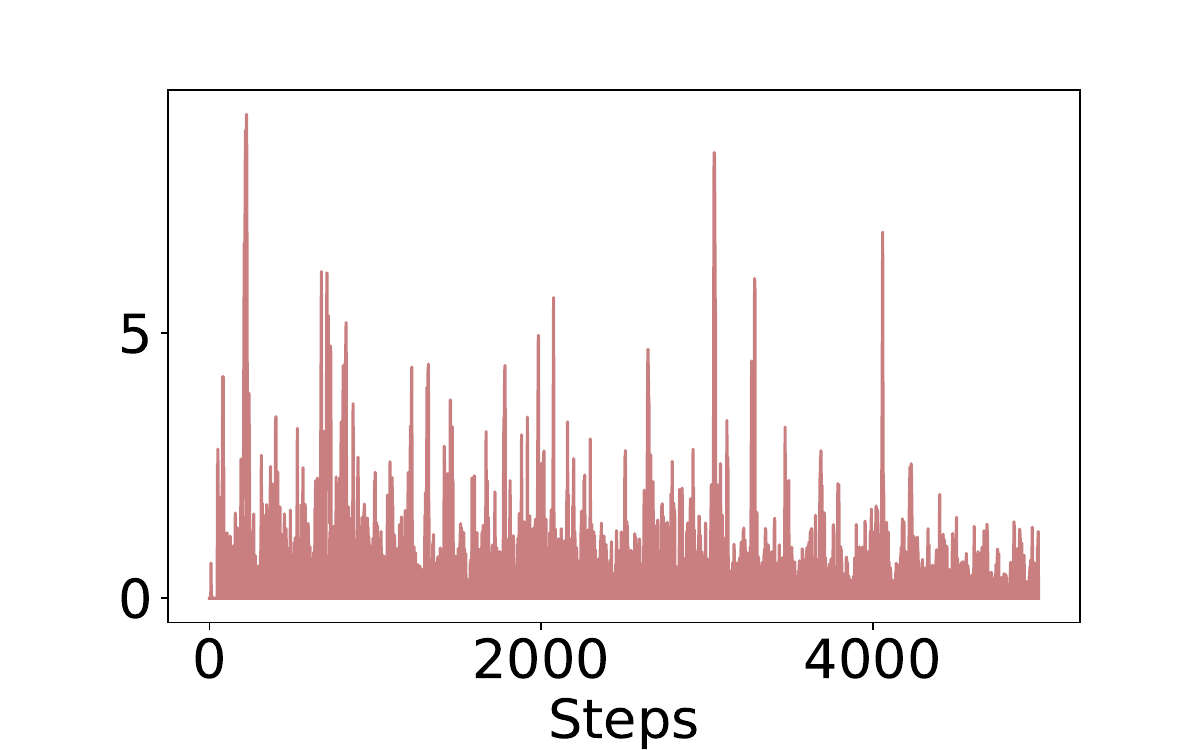}}
		\subfloat[DDPG env\_3]{\includegraphics[width=27mm]{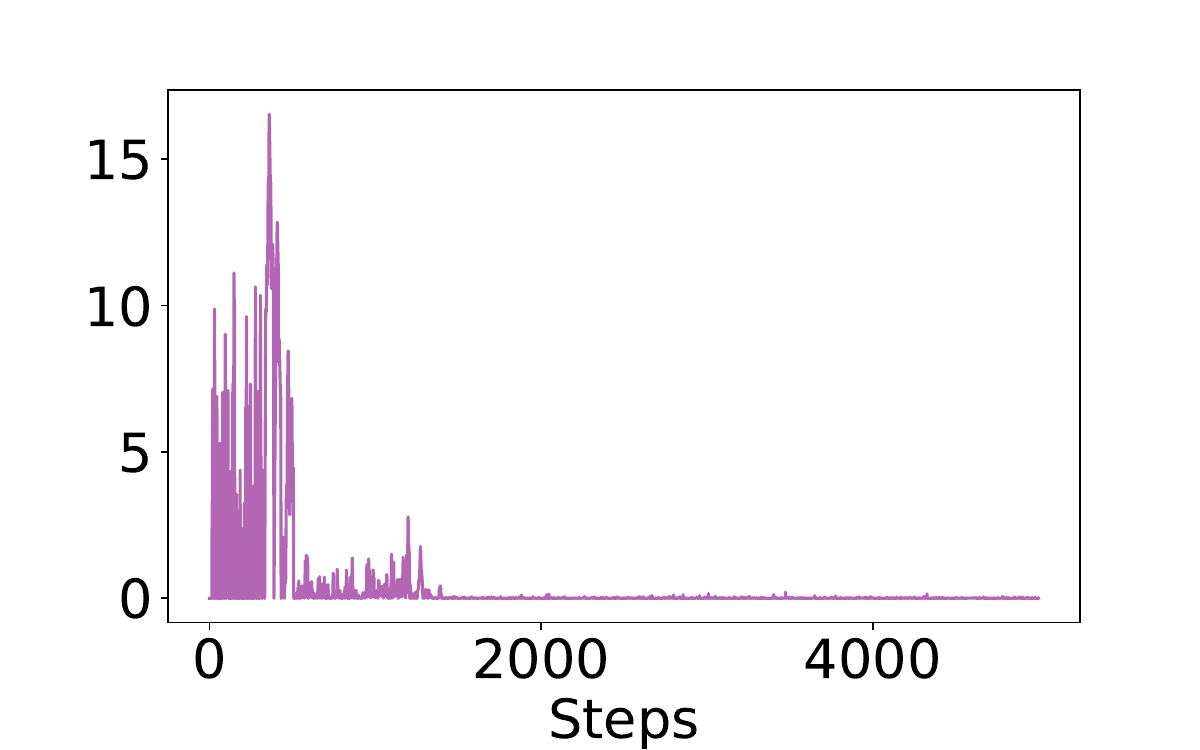}}
		\subfloat[TD3 env\_3]{\includegraphics[width=27mm]{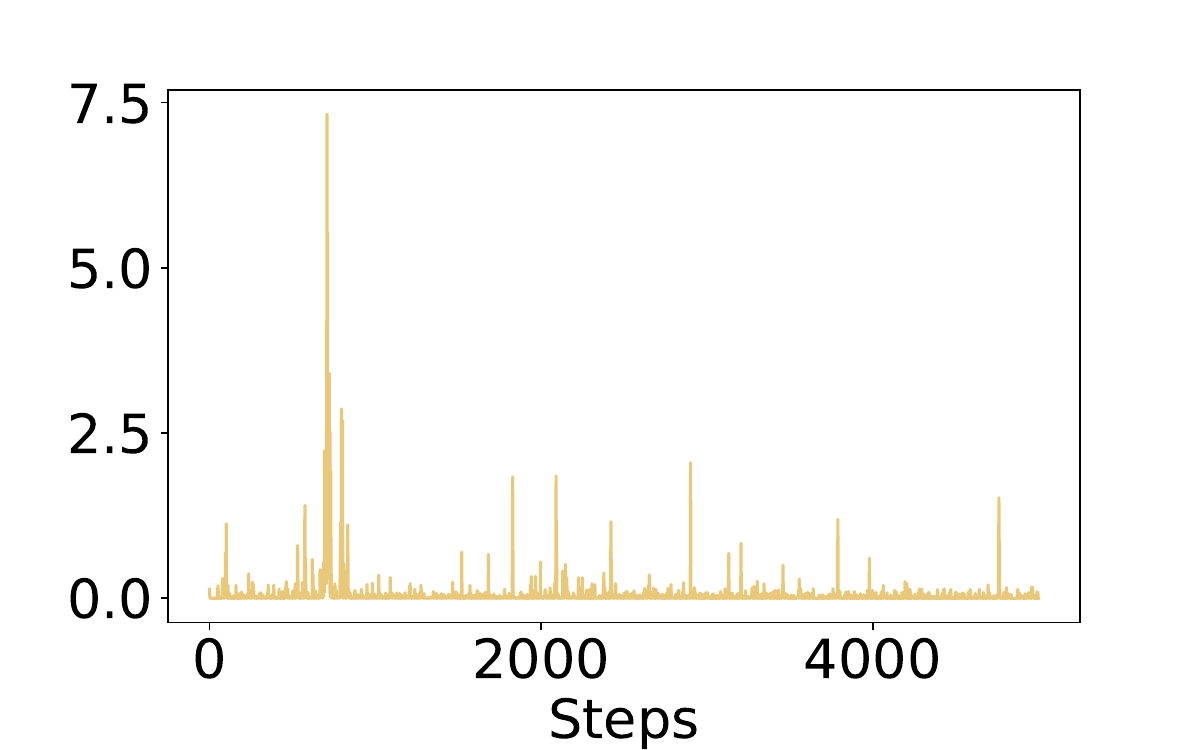}}
		
		\subfloat[PPO env\_4]{\includegraphics[width=27mm]{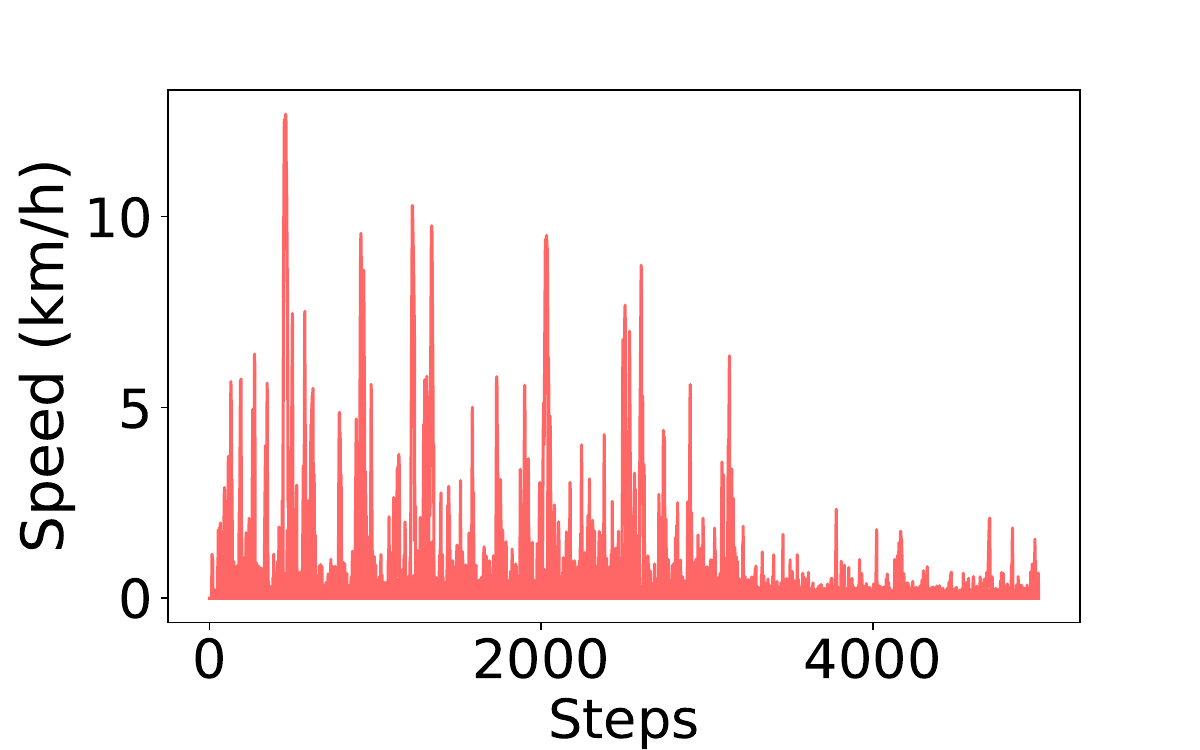}}
		\subfloat[A3C env\_4]{\includegraphics[width=27mm]{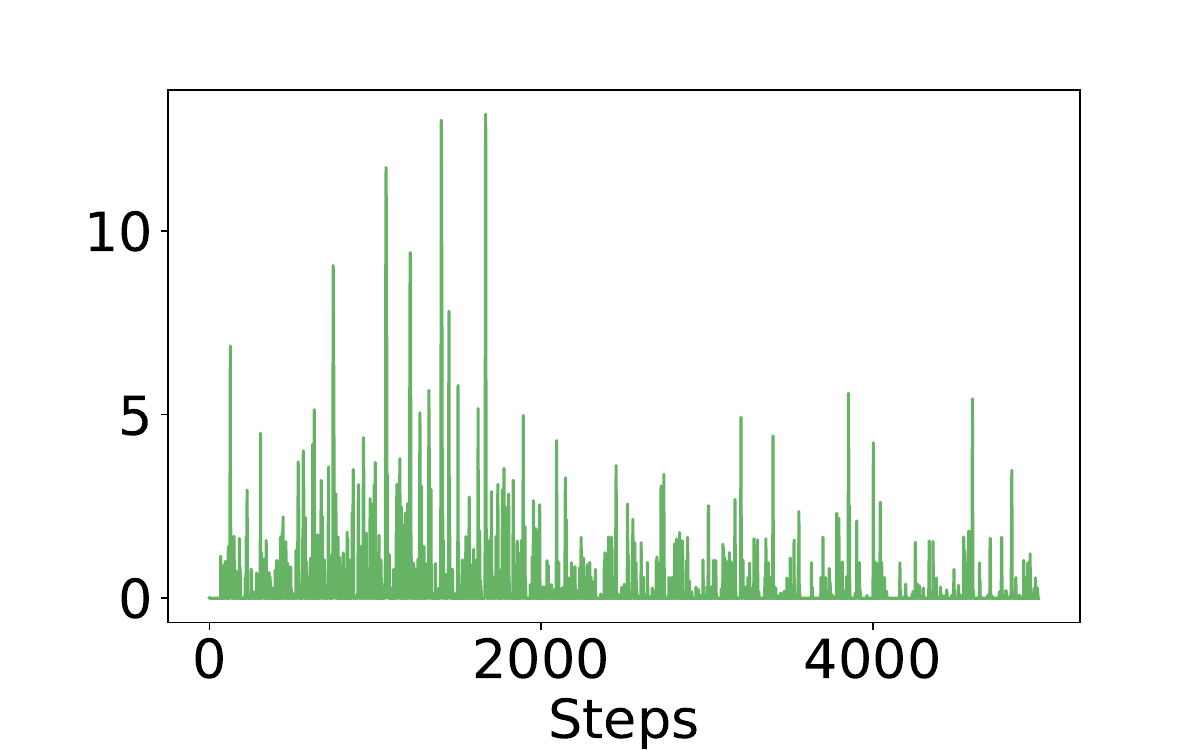}}
		\subfloat[IMPALA env\_4]{\includegraphics[width=27mm]{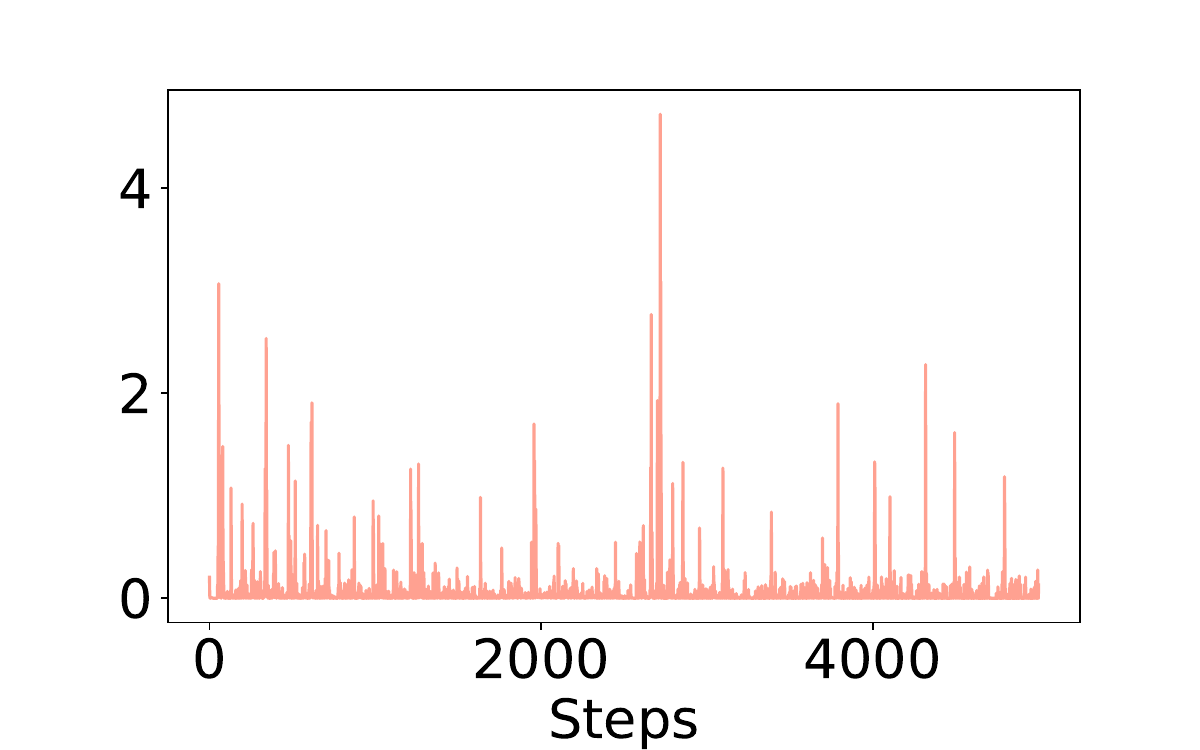}}
		\subfloat[DQN env\_4]{\includegraphics[width=27mm]{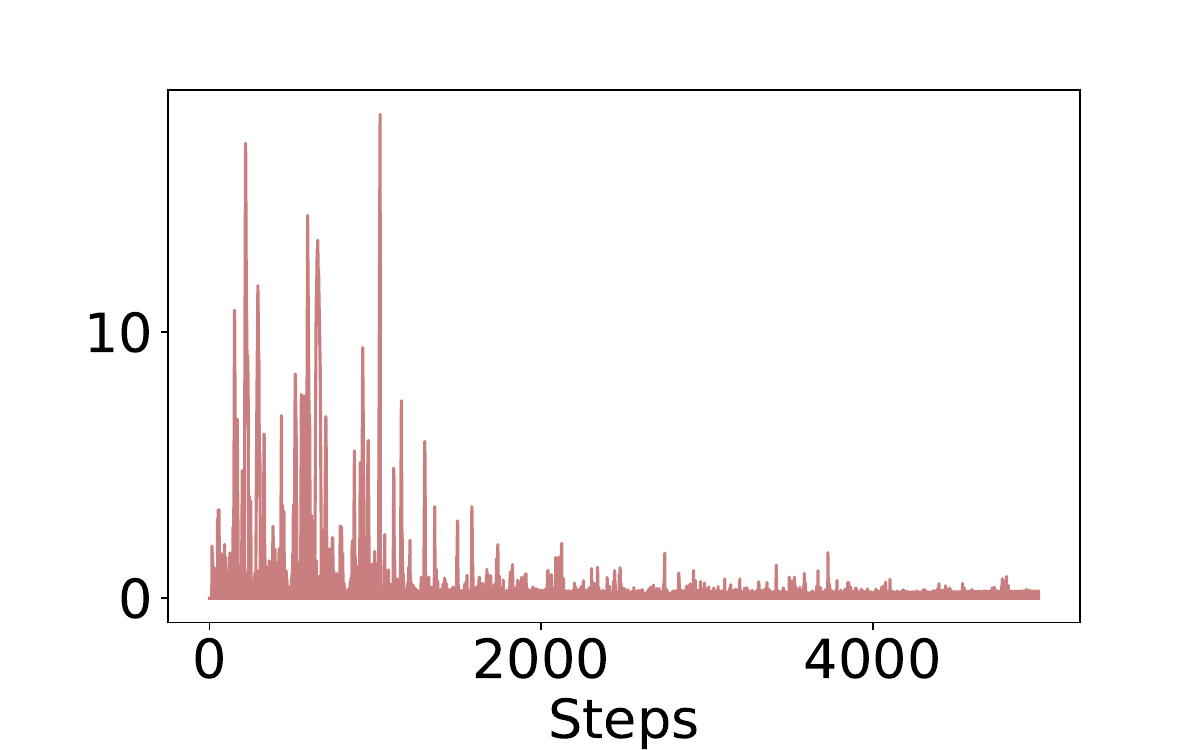}}
		\subfloat[DDPG env\_4]{\includegraphics[width=27mm]{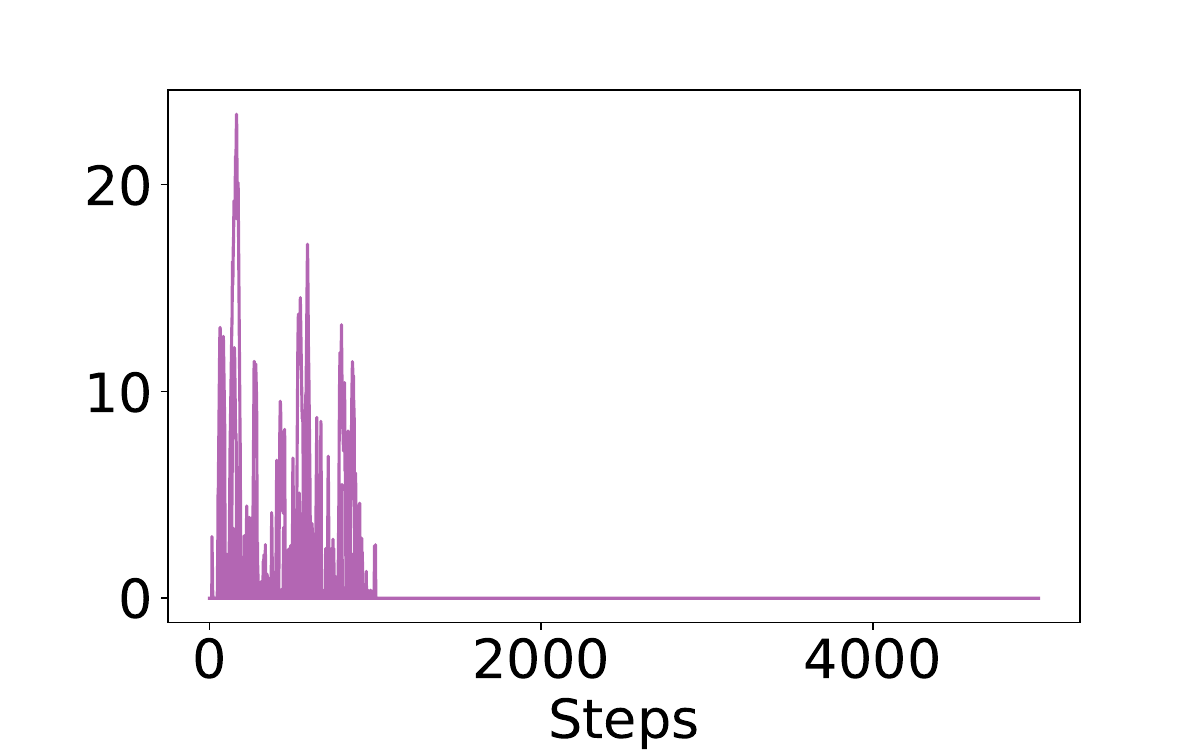}}
		\subfloat[TD3 env\_4]{\includegraphics[width=27mm]{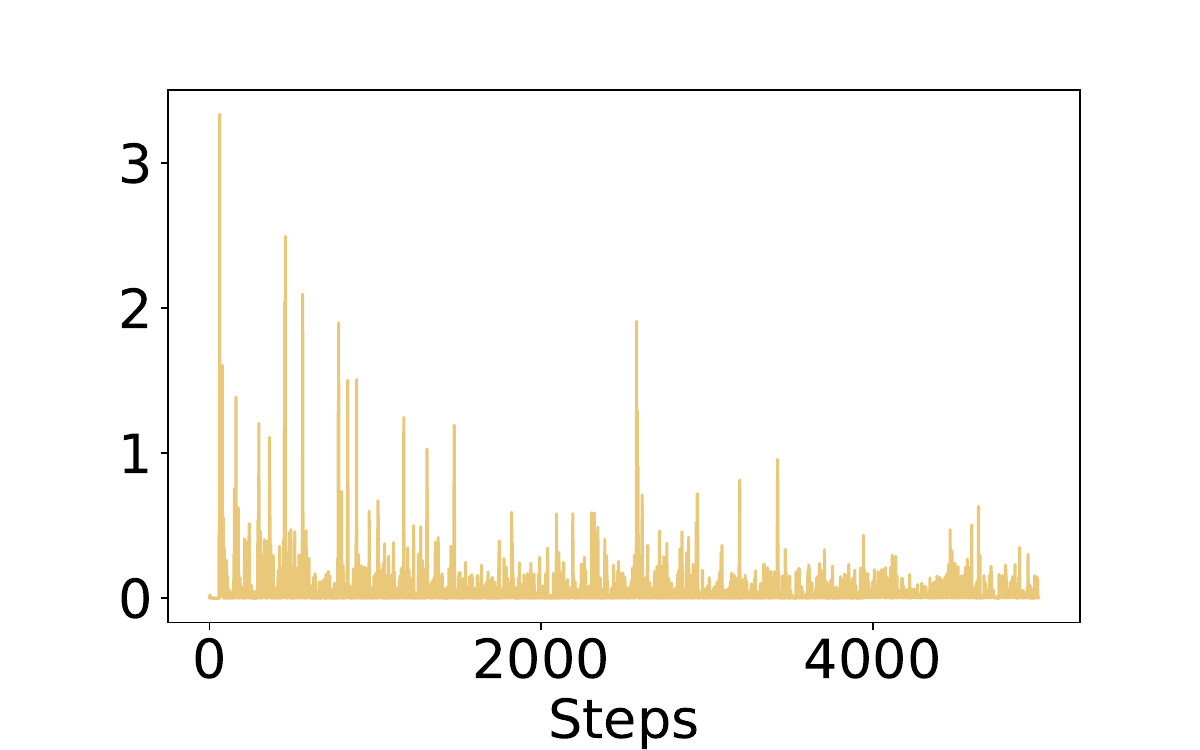}}
		
		\subfloat[PPO env\_5]{\includegraphics[width=27mm]{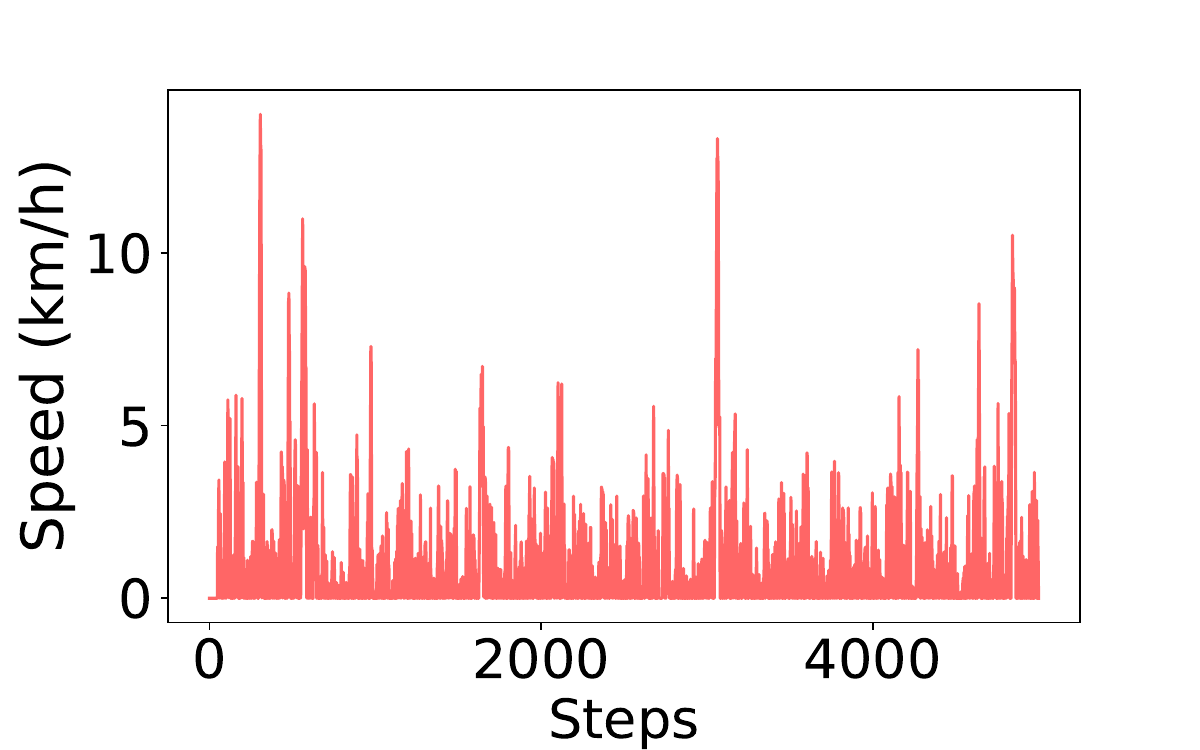}}
		\subfloat[A3C env\_5]{\includegraphics[width=27mm]{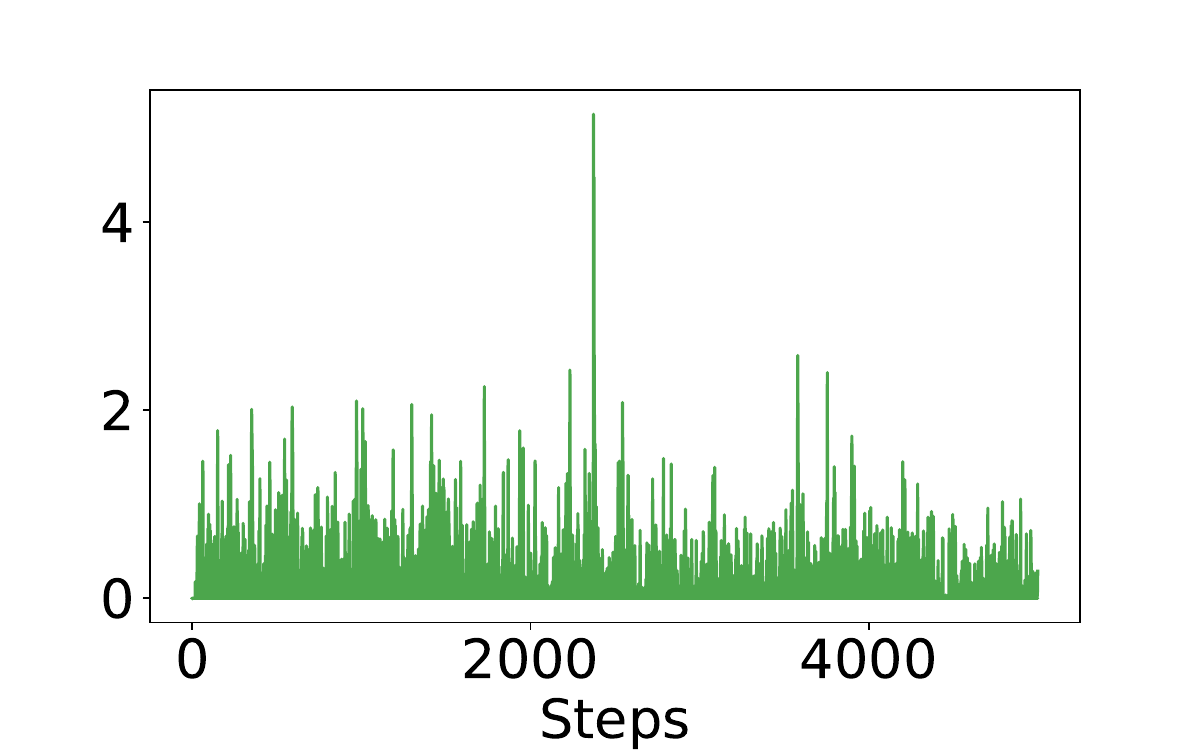}}
		\subfloat[IMPALA env\_5]{\includegraphics[width=27mm]{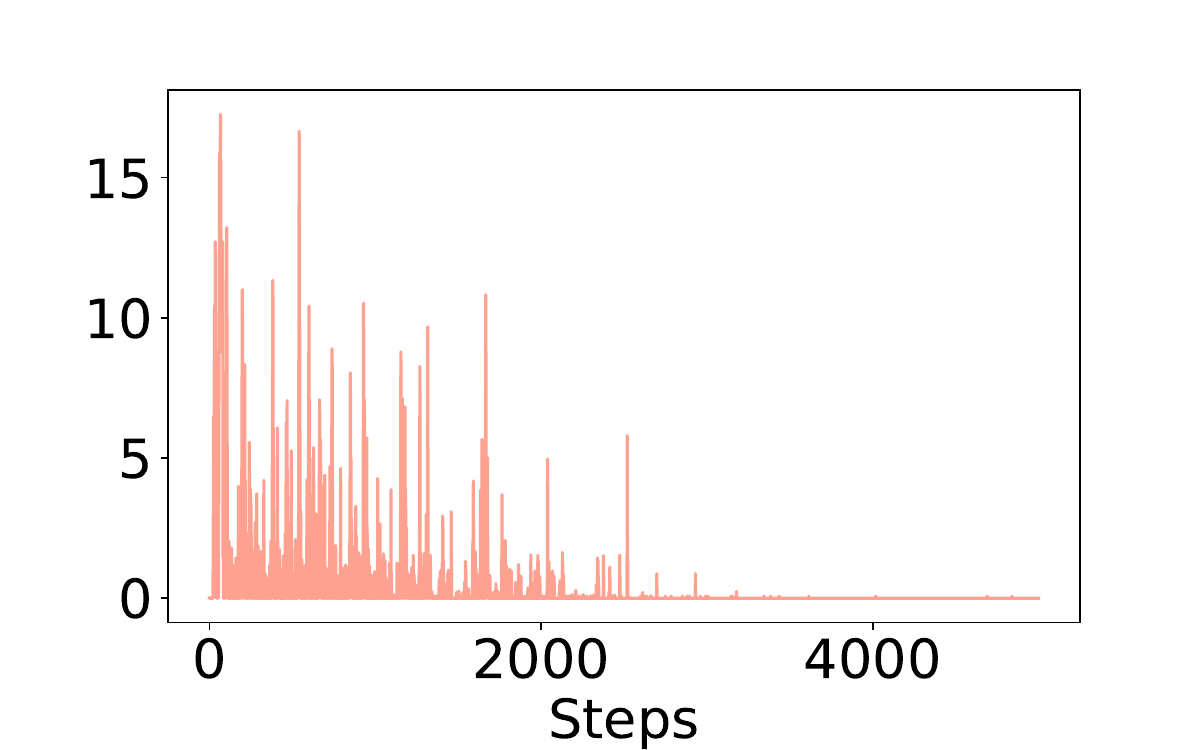}}
		\subfloat[DQN env\_5]{\includegraphics[width=27mm]{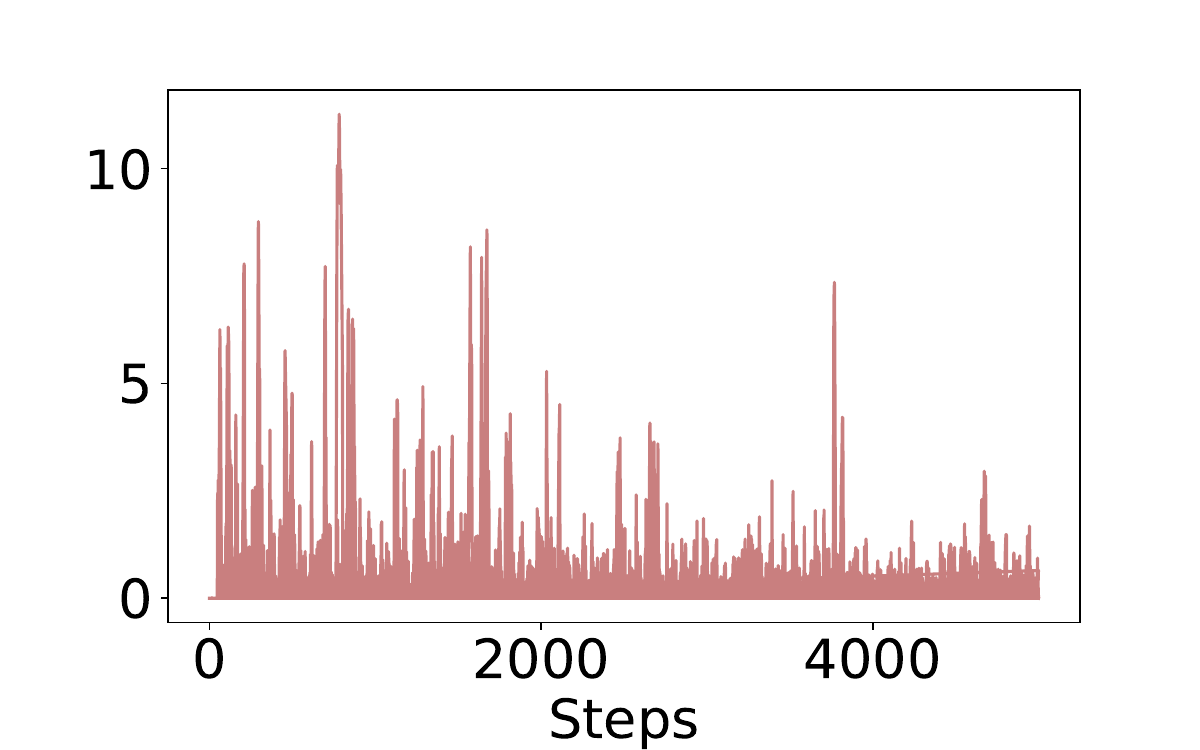}}
		\subfloat[DDPG env\_5]{\includegraphics[width=27mm]{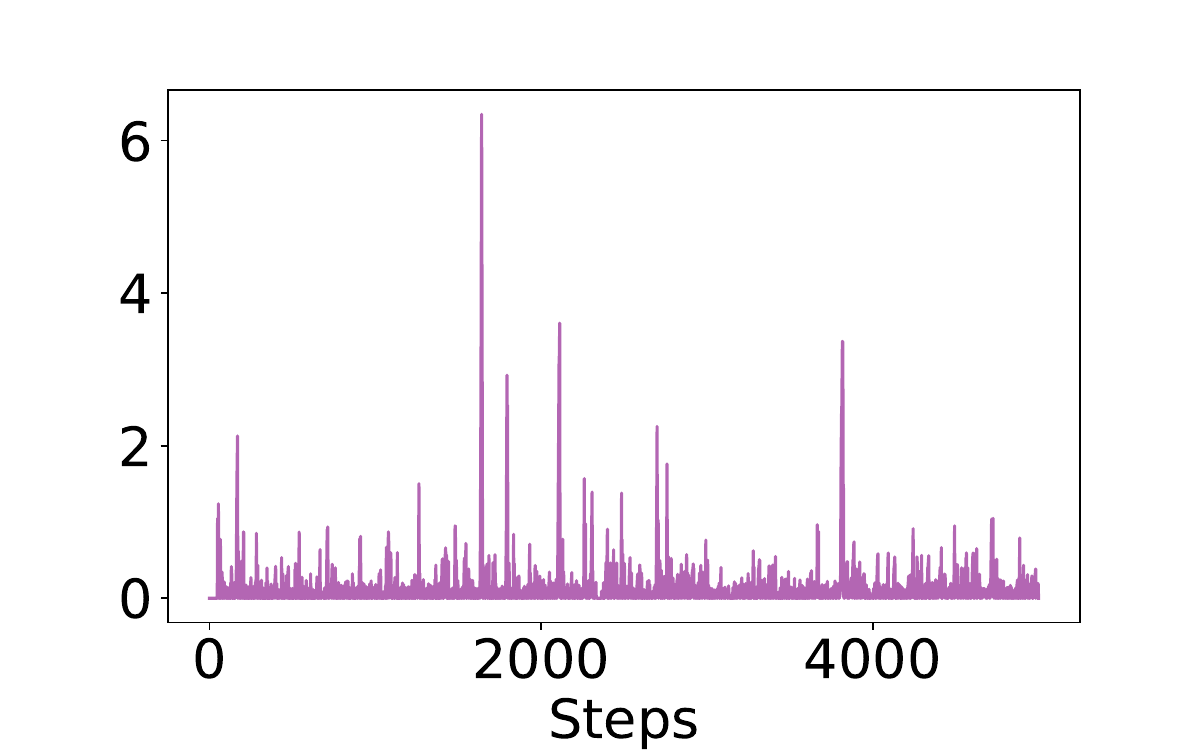}}
		\subfloat[TD3 env\_5]{\includegraphics[width=27mm]{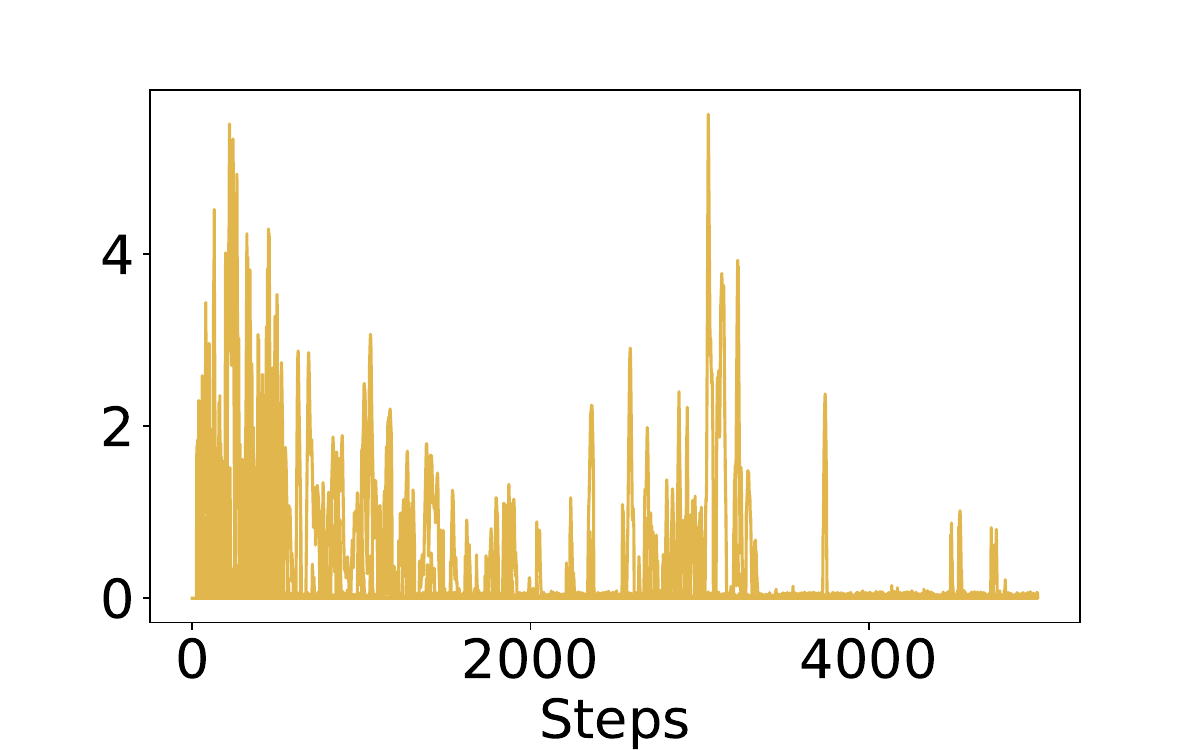}}

		\caption{\label{fig:RQ1Speed}Comparison of the behavior of AC driving agents in terms of forward speed when tested both in a single and multi-agent scenario. Each row represents an average driving speed of DRL AC policies in the driving environments labeled from \textit{env\_1} to \textit{env\_5}.} 
	\end{figure*}

	The evaluation results of the driving performance of DRL-based AC agents are presented in Table~\ref{tab:RQ1} using five Driving Performance Metrics: \textit{CV}, \textit{CO}, \textit{CP}, \textit{OS}, and \textit{TTFC}. AC policies having values closer to 0 are driving error-free, while those near to 1 have a higher failure state. The results of these five metrics should be analyzed jointly with the sixth metric: \textit{SPEED}. This is because a car could have no collision and offroad steering errors due to being stationary. Therefore, we provide Figure~\ref{fig:RQ1Speed}, which shows an AC's driving speed per timestep in a testing simulation. For displaying all the episodic results, we took an average of both \textit{Scenario 1} and \textit{2} for every DRL AC policy across each driving environment setting. The dash sign (-) for the \textit{CV} metric represents Scenario 2 where the evaluation contains only single-agent driving scenarios. Similarly, dash sign (-) for the \textit{CP} metric also represents driving scenarios and environments where pedestrians were not involved. For \textit{TTFC}, we use double dash sign (- -) that represents cases when there is no collision detected by the driving policy.  
	
	\begin{table*}[!htbp]
		\caption{Comparison of the behavior of AC driving agents in terms of \textit{CV}, \textit{CO}, \textit{CP}, \textit{OS} error percentages, and \textit{TTFC} when tested both in a multi (\textit{Scenario 1}) and single-agent (\textit{Scenario 2}) environment. %Collision with cars, collision with road objects, and offroad steering evaluation metrics are represented as \textit{CC}, \textit{CO}, and \textit{OS} respectively.
		}\label{tab:RQ1}
		\begin{center}
			\resizebox{!}{3.9cm}{
				\begin{tabular}{l|l|llllll|llllllll}
					& & \multicolumn{6}{c}{Scenario 1}  & &\multicolumn{6}{c}{Scenario 2} \\
					Environment& Metric & \multicolumn{1}{c}{$\pi_{PPO}$}  &  \multicolumn{1}{c}{$\pi_{A3C}$}  &   \multicolumn{1}{c}{$\pi_{IMPALA}$}
					& \multicolumn{1}{c||}{$\pi_{DQN}$}  & \multicolumn{1}{c}{$\pi_{DDPG}$} 
					& \multicolumn{1}{c|}{$\pi_{TD3}$}
					
					& \multicolumn{1}{c}{$\pi_{PPO}$}  &  \multicolumn{1}{c}{$\pi_{A3C}$}  &   \multicolumn{1}{c}{$\pi_{IMPALA}$}
					& \multicolumn{1}{c||}{$\pi_{DQN}$}  & \multicolumn{1}{c}{$\pi_{DDPG}$} 
					& \multicolumn{1}{c}{$\pi_{TD3}$}
					\\
					\midrule
					\\
					\multirow{5}*{{\textit{env\_1} (Straight)}} &
					\textbf
					{\textit{CV}} & 0.0 & 0.0 &  0.0  & 0.0    &  0.19   &  0.0 & - &  -  &  -&  -&    -&  -  \\
					&	\textbf{\textit{CO}} & 0.76 & 0.0 &  0.97  & 0.0    &  0.19   &  0.1 &  0.2&  0.0  &  0.9 &  0.0 &   0.05 &  0.05 & \\
					&	\textbf{\textit{CP}} & - & - &  -  & -    &  -   &  - &  -&  -  &  -&  -&    -&  - \\
					&	\textbf{\textit{OS}} & 0.9 & 0.0 &  0.96  & 0.7    &  0.9   &  0.2 &  0.2&  0.1  &  0.95 &  0.1 &    0.5 &  0.0 \\  
					&	\textbf{\textit{TTFC} (seconds)} & 268  & - -  &  53.6   & 603    &  67   &  335 &  402&  - - &  120.6&  - - &    60.3&  616.4 \\
					\midrule  
					
					\multirow{5}*{{\textit{env\_2} (Three Way)}} &
					\textbf{\textit{CV}} & 0.0 & 0.0 &  0.0  & 0.17   &  0.3   & 0.0 &  -&  -  &  -&  -&    -&  -  \\
					&	\textbf{\textit{CO}}  & 0.9 & 0.1 &  0.0  & 0.0    &  0.18   &  0.05 &  0.2&  0.006  &  0.19 &  0.0&    0.0&  0.16\\
					&	\textbf{\textit{CP}} & 0.0 & 0.0 &  0.2  & 0.05   &  0.2  &  0. 0&  -&  -  &  -&  -&    -&  - \\
					&	\textbf{\textit{OS}} & 0.2 & 0.1 &  0.6  & 0.3  &  0.9  &  0.0 &  0.0&  0.0  &  0.2 &  0.0&    0.19&  0.39 \\
					&	\textbf{\textit{TTFC} (seconds)} & 241.2 & 636.6 &  268  & 314.9   &  53.6   &  284.08 &  281.4&  670  &  281.4&  - - &    62&  263 \\
					\midrule  
					
					\multirow{5}*{{\textit{env\_3} (Four Way)}} &
					\textbf{\textit{CV}} & 0.1 & 0.1 &  0.1  & 0.0    &  0.0   &  0.0 &  -&  -  &  -&  -&    -&  -  \\
					&	\textbf{\textit{CO}} & 0.0 & 0.0 &  0.7  & 0.26    &  0.0   &  0.0 &  0.0&  0.0  &  0.1&  0.0&    0.0&  0.1 \\
					&	\textbf{\textit{CP}} & 0.1 & 0.0 &  0.1  & 0.05  &  0.1  &  0.0 &  -&  -  &  -&  -&    -&  - \\
					&	\textbf{\textit{OS}} & 0.8 & 0.1 &  0.1  & 0.3  &  0.95  &  0.0 &  0.21&  0.0  &  0.19&  0.49&    0.1&  0.17 \\  
					&	\textbf{\textit{TTFC} (seconds)} & 564.14 & 611.04 &  42.88  & 335   &  248.302   &  - - & - - &  - - &  49 &  - - &  - - &  308.2 \\
					\midrule

					\multirow{5}*{{\textit{env\_4} (Roundabout)}} &
					\textbf{\textit{CV}} & 0.0 & 0.0 &  0.5  & 0.0    &  0.0   & 0.0 &  -&  -  &  -&  -&    -&  -  \\
					&	\textbf{\textit{CO}} & 0.8 & 0.2 &  0.8  & 0.0    &  0.9  &  0.0 &  0.4&  0.0  &  0.05&  0.5&    0.7&  0.59 & \\
					&	\textbf{\textit{CP}} & - & - &  -  & -    &  -   &  - &  -&  -  &  -&  -&    -&  - \\
					&	\textbf{\textit{OS}} & 0.6 & 0.3 &  0.3  & 0.75   &  0.89  &  0.1 &  0.8&  0.08  &  0.6 &  0.64&    0.94&  0.55 \\  
					&	\textbf{\textit{TTFC} (seconds)} & 214.4 & 663.7 &  268.89  & - -  &  53.6   &- - &  335.1& - - & 660 &  26.8&    51&  120.6 \\
					\midrule  
					
					\multirow{5}*{{\textit{env\_5} (Merge)}} &
					\textbf{\textit{CV}} & 0.0 & 0.0 &  0.0  & 0.0    &  0.0   &  0.0 &  -&  -  &  -&  -&    -&  -  \\
					&	\textbf{\textit{CO}} & 0.9 & 0.0 &  0.0  & 0.0   &  0.0   &  0.05 &  0.2&  0.0  &  0.0&  0.0&    0.34&  0.05 \\
					&	\textbf{\textit{CP}} & - & - &  -  & -    &  -   &  - &  -&  -  &  -&  -&    -&  - \\
					&	\textbf{\textit{OS}} & 0.5 & 0.1 &  0.4 & 0.8   &  0.98  &  0.6 &  0.8&  0.0  &  0.06&  0.1&    0.96&  0.21 \\  
					&	\textbf{\textit{TTFC} (seconds)} & 323.6 & - - & - -  &- -  &  - - &  634.22 &  536&  - -  &  - - & - -&    33.5&  650 \\
					\midrule  
				\end{tabular}
			}		
		\end{center}
	\end{table*}

\subsubsection{{$\pi_{PPO}$}}
$\pi_{PPO}$ based AC policy was able to avoid collision with other vehicles except for some minimal collision states in \textit{env\_3}. On the other hand, $\pi_{PPO}$ performed poorly in respect of road and pedestrian collisions throughout the multi-agent driving scenarios. From simpler environments as \textit{env\_1} to much more complex driving conditions as \textit{env\_4} and env\_5, we see many offroad steering errors made by the $\pi_{PPO}$ algorithm. We see similar behavior in the single-agent scenario, where it performs no offroad steering errors in only \textit{env\_3}, while the agent collides with other road objects and drives off-lane throughout the driving settings. PPO-based AC policy also ends up having its first collisions in early timesteps of the episodes, whether tested in multi-agent or single-agent scenarios. $\pi_{PPO}$ performance with respect to $SPEED$ has been mostly consistent regardless of the collisions and offroad steering states as stated above.

\subsubsection{\textbf{$\pi_{A3C}$}}

From Table~\ref{tab:RQ1}, it is clear that A3C algorithm-based policy $\pi_{A3C}$ performs best with no collision with other vehicles in four out of five environments. It also performs with minimal road collisions and offroad steering errors across all driving environments. The driving behavior remains similar in both multi and single-agent testing scenarios. While testing against pedestrians within \textit{env\_2} and \textit{env\_3}, it successfully avoids any collision. $\pi_{A3C}$ only gets very minor states of collision with other vehicles in \textit{env\_3} multi-agent scenarios. $\pi_{A3C}$ policy while facing these minor collisions was detected in the first half of the episodes, but the driving agent was able to recover from the failure state. The time to detect the first collision in such scenarios is also comparatively longer than other driving policies as shown in the Table. In terms of $SPEED$, the A3C algorithm-based policy $\pi_{A3C}$ keeps a decent driving speed throughout episodic timesteps. A3C overall performs consistently throughout the testing episodes with both high and low-speed values.

\subsubsection{\textbf{$\pi_{IMPALA}$}}
$\pi_{IMPALA}$ can be observed in our experiments as the weakest discrete action space algorithm. In a multi-agent scenario, it sometimes avoids collisions with other driving vehicles, but most of the time ends up colliding with other road objects and pedestrians with a large percentage of offroad steering errors. The driving behavior is slightly improved when driving alone within \textit{env\_4} and env\_5. Other than single-agent testing, $\pi_{IMPALA}$-based AC policy in \textit{env\_4} is also found in its first collision state in the earlier timesteps of the episodes. This can also be visualized in terms of $SPEED$, where the AC policy performs its best in \textit{env\_4}, while in the rest it drastically decreases its speed after early collision states.

\subsubsection{\textbf{$\pi_{DQN}$}}
$\pi_{DQN}$ overall performs second best within discrete action space algorithms. In 4 out of 5 environments, the $\pi_{DQN}$ AC policy avoids vehicle collisions in the multi-agent scenario. However, it constantly drives offroad in multi-agent conditions and also collides a few times with pedestrians while facing them in \textit{env\_2} and \textit{env\_3}. The driving performance of $\pi_{DQN}$ improves a lot when driving alone in single-agent scenarios across all environments. It only faces road object collision states in a difficult and complex \textit{env\_4} environment. The driving policy even after getting its first collision in the early stage of the episodes recovers to a better driving state. The $SPEED$ factor is maintained quite well by $\pi_{DQN}$ AC policy throughout the testing phase.

\subsubsection{\textbf{$\pi_{DDPG}$}}
In continuous action space algorithms, we first evaluate the robustness of $\pi_{DDPG}$ across various driving conditions. The $\pi_{DDPG}$ policy performs the worst in overall driving while following the road lane and therefore gets into the majority of the offroad driving states whether it is a single or multi-agent scenario. It also collides with pedestrians while testing in \textit{env\_2} and \textit{env\_3}. The time to detect the first collision is also earlier in four out of 5 driving environments whether it is a first or second scenario. The driving $SPEED$ of $\pi_{DDPG}$ can be interpreted as similar to $\pi_{IMPALA}$ where it slows down in early timesteps of the episodes due to collisions and offroad steering errors.

\subsubsection{\textbf{$\pi_{TD3}$}}
In continuous action space algorithms, $\pi_{TD3}$ on the other hand makes a lot better driving decisions as compared to $\pi_{DDPG}$. It successfully avoids collision with other vehicles within multi-agent scenario settings and also gets in no contact with pedestrians across \textit{env\_2} and \textit{env\_3} driving conditions. The $\pi_{TD3}$ AC policy faces some minor road collision and offroad steering states in multi-agent scenario testing. The driving behavior slightly drops when $\pi_{TD3}$ is tested in single-agent scenarios, especially in \textit{env\_4} where both offroad steering and road collision percentages increases. In the case of environments where it had any collision, the time to detect the first one is far greater than the ones detected in the DDPG AC policy. The forward $SPEED$ conditions of $\pi_{TD3}$ is in general slower than $\pi_{DDPG}$, and therefore able to drive mostly safer in comparison to $\pi_{DDPG}$ from start to end.  

In summary, the experimental results of testing AC policies in a single and multi-agent scenario indicate that the A3C and TD3-based driving agents perform better than the rest of the DRL agents in both single and multi-agent scenarios within five driving environment settings, answering RQ1. 
	
 \subsection{RQ2: \textbf{Success rate of AC driving policies in achieving multiple driving objectives}}\label{sec:RQ2}
\subsubsection{\textbf{Safety}}

\begin{figure}[!htbp]
	\centering
	
	\includegraphics[width = 0.48\textwidth,height=0.1\textheight]{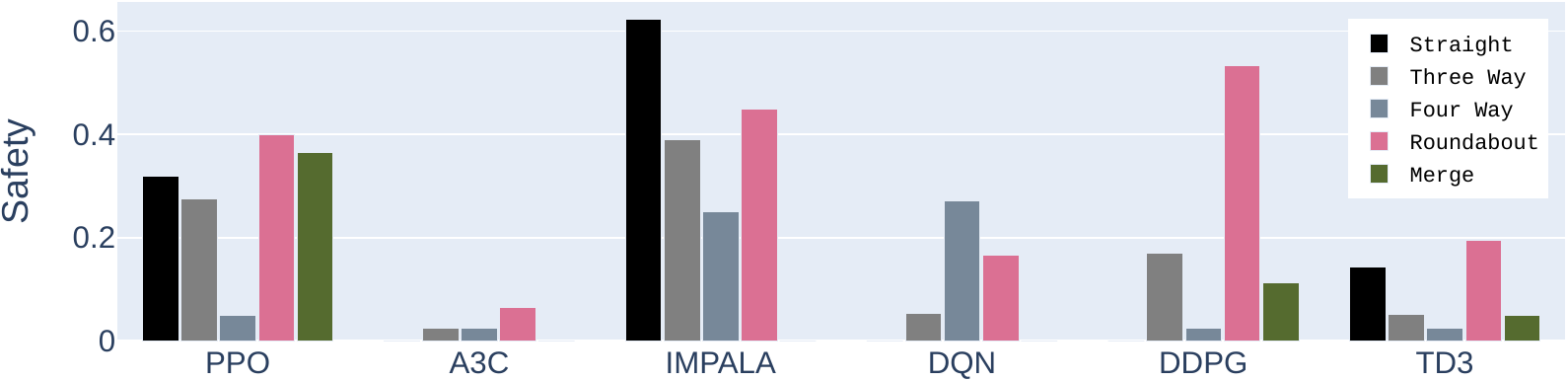}
	\caption{Illustration of the success rate within DRL AC policies with respect to safety. Each algorithm in discrete and continuous action space is evaluated across all driving environments.}
	\label{fig:Safety}
\end{figure}

By visualizing the results in Figure~\ref{fig:Safety}, we can see the average success rate of each DRL algorithm in achieving safety across all driving environments. A value closer to zero is considered much safer than the ones closer to one displayed in the figure. As thoroughly discussed in Section~\ref{sec:RQ1} we can clearly see that $\pi_{A3C}$ overall performs the best in terms of avoiding collisions with drivers, pedestrians, and other road objects. $\pi_{DQN}$ performs similarly to $\pi_{A3C}$ within \textit{env\_1} and env\_5 but performs less safely in the rest of the 3 environments. $\pi_{PPO}$ other than \textit{env\_3} has a mediocre success rate in terms of safety, whereas $\pi_{IMPALA}$ other than \textit{env\_5} performs the worst overall. In continuous action space, $\pi_{TD3}$ shows the overall safest driving behavior across all environmental conditions as compared to $\pi_{DDPG}$.

\subsubsection{\textbf{Efficiency}}

\begin{figure}[!htbp]
	\centering
	
	\includegraphics[width = 0.48\textwidth,height=0.1\textheight]{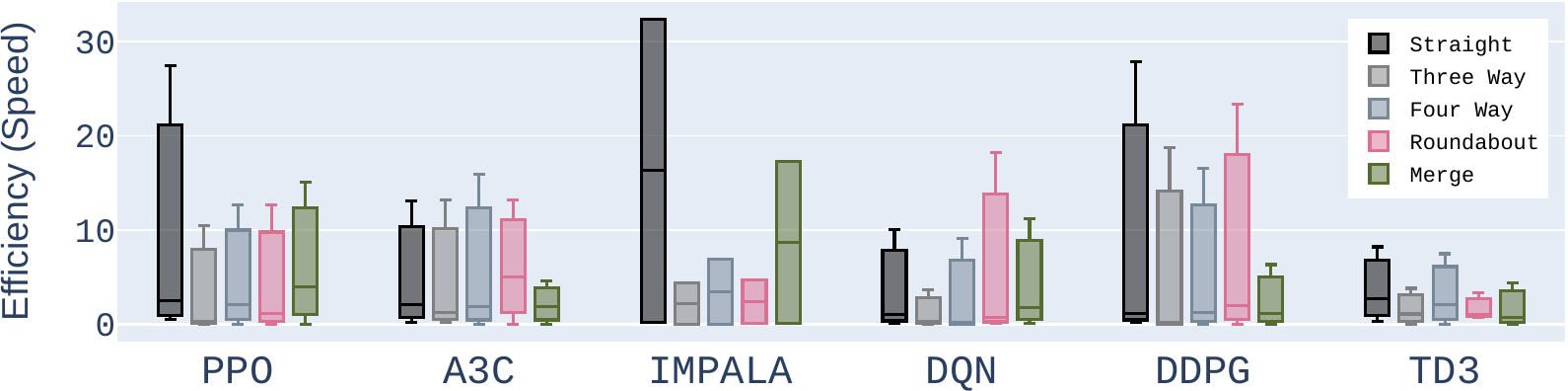}
	\caption{Illustration of the success rate within DRL AC policies with respect to efficiency in speed. Each algorithm in discrete and continuous action space is evaluated across all driving environments.}
	\label{fig:Efficiency_Speed}
\end{figure}

When it comes to the success rate for measuring efficiency, we need to analyze the results with respect to both the driving $SPEED$ and $DISTANCE$ covered for each DRL AC policy. Figure~\ref{fig:Efficiency_Speed} displays the box plot of minimum, maximum, and median speed values throughout the testing phase. In discrete action space, $\pi_{A3C}$ maintains consistent speed across all environments, whereas in continuous space algorithms, $\pi_{TD3}$ is much slower than $\pi_{DDPG}$. 

\begin{figure}[!htbp]
	\centering
	
	\includegraphics[width = 0.48\textwidth,height=0.1\textheight]{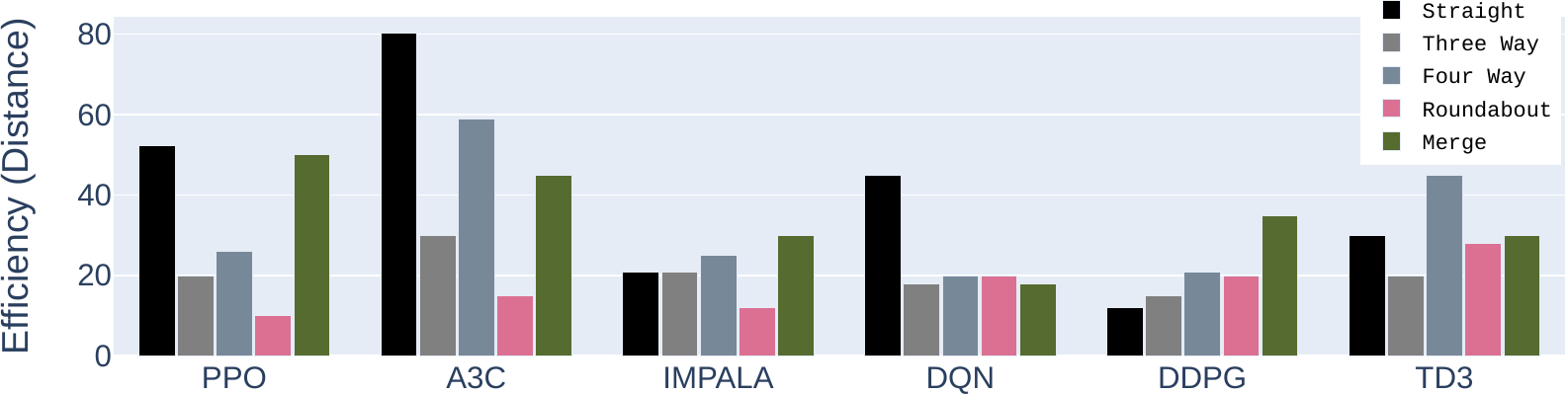}
	\caption{Illustration of the success rate within DRL AC policies with respect to efficiency in covering distance. Each algorithm in discrete and continuous action space is evaluated across all driving environments.}
	\label{fig:Efficiency_Distance}
\end{figure}

To fully understand the efficiency, we also visualize Figure~\ref{fig:Efficiency_Distance} to measure the average distance covered by each DRL AC policy, since speed does not alone guarantee the efficiency of AC policies. $\pi_{A3C}$ overall covers the majority of the distance across driving conditions. A similar pattern can be observed for $\pi_{TD3}$ in terms of driving more distance than $\pi_{DDPG}$ by also maintaining a slower speed as discussed above. 

 \subsubsection{\textbf{Lane Keeping}}
As described in Section~\ref{sec:Reward}, one of the objectives of DRL AC policies in this experiment is to also focus on lane-keeping behavior. A safer car is much more likely to drive within a lane with minimum offroad steering errors, but these are two different goals that separately need to be achieved. In terms of evaluating success rate, $\pi_{A3C}$ once again performs the best. Other than \textit{env\_4} it learns to fully drive in its lane throughout the experiments. $\pi_{IMPALA}$ remains inconsistent with its lane-keeping decisions since it performs the worst in simplistic environments such as \textit{env\_1} but performs better in complex conditions such as \textit{env\_4}. $\pi_{PPO}$ and $\pi_{DQN}$ are somewhat similar in lane-keeping decision-making where both do not perform well in \textit{env\_4}. When coming towards continuous action space, $\pi_{TD3}$ performs better in terms of driving within its lane in all of the environments compared to $\pi_{DDPG}$ by a large margin.
\begin{figure}[!htbp]
	\centering
	
	\includegraphics[width = 0.48\textwidth,height=0.1\textheight]{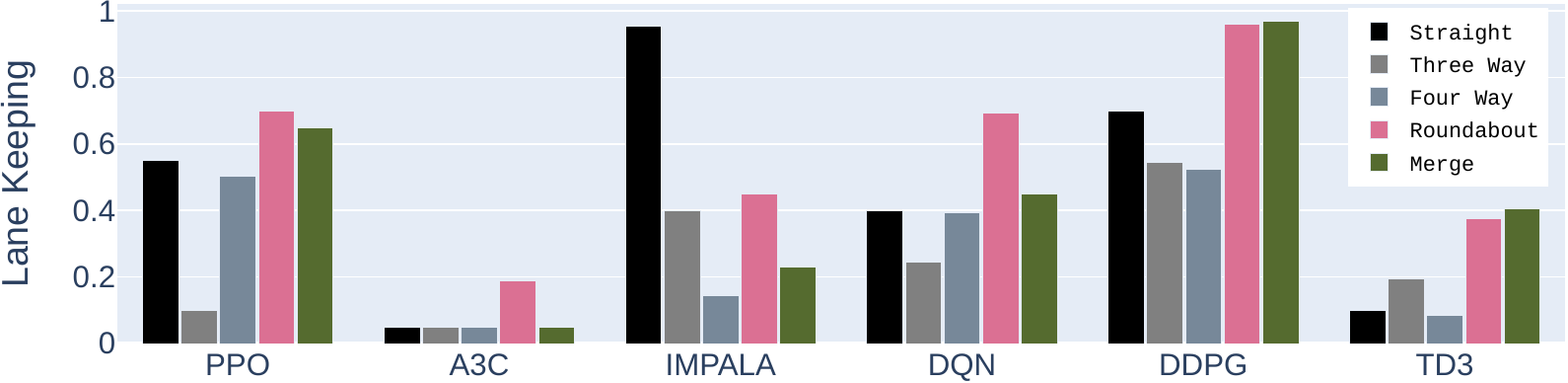}
	\caption{Illustration of the success rate within DRL AC policies with respect to lane keeping. Each algorithm in discrete and continuous action space is evaluated across all driving environments}
	\label{fig:Lane_Keeping}
\end{figure}

In summary, the analysis of the success rate of testing AC policies in RQ2 in all of the environments indicates that the A3C and TD3-based driving agents perform safe driving decisions, cover the majority of the distance, and perform minimal offroad steering errors. The analysis also shows that $\pi_{A3C}$ in discrete action space learns to drive with consistent speed while $\pi_{TD3}$ tends to keep lower speed as a trade-off in order to make better driving decisions.

\section{Discussion and Open Research Directions}\label{sec:Discussion}
This section provides our observations and thoughts on the effectiveness and robustness of the six evaluated DRL algorithms for their use in AD research, specifically, as AC agents used for testing the driving agents. 

\textbf{PPO} as AC, in general, did not perform better than the rest of the algorithms while driving in a single or multi-agent scenario. Standard PPO policies are often stuck at non-optimal actions while learning since they are very sensitive to sparse high rewards. However, the PPO-based AC policy drove efficiently with a good driving speed and covered a lot of distance. This shows that PPO-based policies need to focus more on safety and lane-keeping driving behaviors as per our experiments.

\textbf{A3C} as AC performs the best in both single-agent and multi-agent environments. With its default implementation, the critic model seems to learn the value function well while actors are independently learning their driving policies in parallel. A3C seems to work well where training is performed on input camera images~\cite{A3C}\cite{A3C2}\cite{A3C3} as in our case. One of the future directions is to investigate how A3C-based policy would perform if it tries to drive with more speed than the rest of discrete action space policies.
	
\textbf{IMPALA} has shown great results when used before in the same simulation environment~\cite{R52}, but the main difference in that work is that the authors implemented the IMPALA algorithm for training multi-agent ACs using shared connected AC weights. In our work, we train a non-shared multi-agent AC which shows to not only fail in single-agent testing but also in a multi-agent competitive scenario. The algorithm is unable to effectively use the trajectories gathered from actors that are decoupled from the learner. IMPALA algorithm has only been successful in driving in complex driving conditions such as roundabouts while also performing fewer offroad steering mistakes. This shows that the algorithm has the potential to further tune the driving capability in complex urban driving scenarios.

Default \textbf{DQN} is the most used algorithm within the DRL community and even in AD research, even though training ACs using DQN in our case did not achieve good results as compared to A3C. A lot of extensions have been proposed since the popularity of DQN in atari games. Therefore, one prosperous research direction is to explore the continuous action space as in DDPG~\cite{Lillicrap2016ContinuousCW}.  

\textbf{DDPG} is widely used for training AC  policies~\cite{DDPG}\cite{DDPG2}\cite{DDPG3}, and we show that the improved DRL model \textbf{TD3} performs better during single and multi-agent testing scenarios. The algorithmic advantages of the TD3 policy over DDPG really help in learning a multi-agent driving policy. At the same time, TD3 maintained a very low speed in our experiments to perform safe and efficient driving. Just like A3C in discrete action space, TD3 in continuous action space needs to be further looked into with respect to higher speed and analyze the safety and lane keeping performing in parallel.

	\section{Threats to Validity}\label{sec:Threat}
	
	\textbf{Non stationary multi-agent driving environments}.
	In multi-agent non-stationary environments, each agent's transition probability and reward function depends on the actions of all the agents since they change every time with the actions performed by the agents. DRL research for AD is mainly focused on driving in a single-agent stationary MDP environment. Driving behavior is affected a lot when tested in a multi-agent scenario due to the non-stationary driving environment~\cite{Papoudakis2019DealingWN}. This is one of the key threats to the existing DRL-based AD research that is performed only in a single-agent scenario. As explained in Section~\ref{sec:RQ1}, the driving performance of AC agents is affected significantly when they are exposed to more AC agents in a multi-agent setting.

	%\textbf{Choice of driving scenarios}.
	%As mentioned in Section~\ref{sec:env}, we are using \textit{four-way intersection} and \textit{T-intersection} for testing AC agents in a single as well as a multi-agent scenario. The primary reason for selecting such scenarios is because they represent high-complexity driving scenarios, by facing cars from various directions in an intersection. However, the results are collected in the mentioned driving environments only. Our future research direction is to further investigate the driving performance of each AC and adversarial agent using different driving scenarios.
	
	\textbf{Choice of DRL algorithms}.  
	We have selected a total of six DRL algorithms to test the robustness of AC driving agents as described in Section~\ref{sec:DeepRL}. While there could be other suitable DRL algorithms, we have selected ours to cover a range of DRL categories, including value-based, policy-based, and actor-critic-based. Furthermore, our benchmarking framework relies on the Ray RLlib framework~\cite{R54} which makes it possible to implement a competitive DRL-based multi-agent driving environment. At the moment, the choice of DRL algorithms also partly depends on which DRL algorithms work smoothly while integrated with Ray RLlib, Tensorflow~\cite{R53}, and CARLA-specific versions~\cite{R51}.  
	%\subsection{Lack of support for continuous action space algorithms}

 \textbf{Hyperparameters tuning}.
Hyperparameter tuning is considered a very sensitive part of training DRL policies, and to mitigate this threat, we have used the best hyperparameters reported in the past implementations in literature as a starting point, while also utilizing a hybrid hyperparameter tuning algorithm to search for the most optimum hyperparameters. Experimenting with hyperparameter tuning can be tested further in future AD research.
	\section{Conclusion}\label{sec:Conclusion}
	
	In this work, we compare the robustness of DRL-based policies while training ACs agents. 
	By first training DRL AC policies in a multi-agent environment, we test their driving performance in both single and multi-agent scenarios. We analyze the robustness of AC policies using six evaluation metrics within both multi and single-agent scenarios. We also propose a comprehensive multi-objective reward function designed for the evaluation of DRL AC driving agents. Based on our proposed reward function, we analyze the overall success rate in achieving safety, efficiency, and lane keeping for each DRL policy as described in Section~\ref{sec:Results}.

	\bibliographystyle{IEEEtran}
	\bibliography{refs}

\end{document}